# Optimal any-angle path planning in static and dynamic environments

Yiyuan Zou *, Clark Borst

*Control and Simulation, Faculty of Aerospace Engineering, Delft University of Technology, Delft, The Netherlands*

ABSTRACT

Any-angle path planning extends traditional graph-based path planning by allowing movement between any pair of vertices, rather than being restricted by predefined edges. It can find straighter and shorter paths in continuous space with graphs, making it particularly suitable for navigation in open areas. Many any-angle path-planning algorithms have been proposed, but only a few can guarantee optimal solutions. To address this challenge, this article focuses on optimal any-angle path planning on grids and introduces two general techniques that accelerate computation while preserving optimality in both static and dynamic environments (where dynamic environments refer to those with moving obstacles in this article): 1) elliptical forward expansion, which leverages ellipse-based neighborhoods to restrict the search space, and 2) field of view, which replaces traditional line-of-sight methods to speed up visibility checks. To integrate these two techniques, inverted and forward scanning are introduced, denoted as "-i" and "-f" for the corresponding algorithmic variants. Inverted scanning establishes visual connections from open nodes, whereas forward scanning initiates scans from closed nodes. Building on the proposed techniques, Zeta* and Zeta*-SIPP are developed for static and dynamic environments respectively. Zeta*, when combined with forward scanning (i.e., Zeta*-f), is similar to the state-of-the-art algorithm Anya and attains comparable performance. Unlike Anya, Zeta* can be readily extended to other settings. For example, Zeta*-SIPP, which extends Zeta* by integrating Safe Interval Path Planning (SIPP), can find time-optimal any-angle paths in dynamic environments. Benchmarking results show that Zeta*-SIPP achieves more than an order-of-magnitude speedup over the state-of-the-art optimal planner TO-AA-SIPP when using either scanning method (i.e., Zeta*-SIPP-i or Zeta*-SIPP-f). Overall, this research identifies the key requirements for achieving optimal any-angle path planning and introduces a unified approach suitable for different environments.

## 1. Introduction

Path planning is a fundamental problem across various domains, including aerospace, transportation, robotics, and computer games. It typically involves finding an optimal path between two points in a given space, while avoiding collisions with static obstacles and, if present, dynamic obstacles. Over the years, numerous algorithms have been developed to address this problem under a wide range of real-world conditions. Among them, A* [1] stands out as one of the most classic and widely recognized algorithms. However, as a graph-based algorithm, traditional A* is highly affected by the structure of the predefined graph. On regular square grids, A* generally considers only the eight adjacent neighbors of the current node during the search, restricting its movement to 45-degree increments at each step, as shown in Fig. 1a.

* Corresponding author.
*E-mail addresses:* y.zou@tudelft.nl (Y. Zou), c.borst@tudelft.nl (C. Borst).

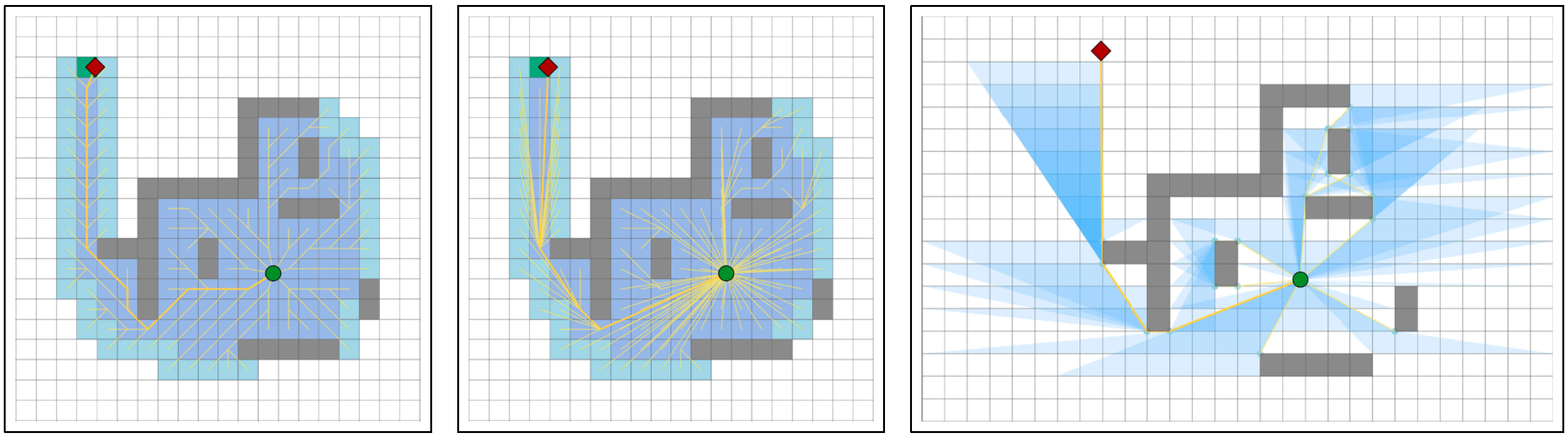

(a) A* (b) Theta* (c) Anya

**Fig. 1.** Search trees (yellow lines) and explored nodes (blue grids or regions) of A*, Theta* and Anya[1]. The green dot is the start point and the red diamond is the target point. The dark gray cells represent static obstacles.

To address this issue, some post-hoc smoothing techniques can be applied to straighten the final generated paths [3]. However, they do not guarantee the discovery of true shortest paths and may be difficult to yield effective results in complex environments. Any-angle path planning has thus been proposed [4], which allows any-angle turns at graph vertices rather than being limited to fixed graph edges. Theta* [5] is one of the most well-known any-angle path-planning algorithms. As shown in Fig. 1b, Theta* generates a more direct and straighter search tree than A*, resulting in a shorter final path. Technically, any-angle path planning integrates path smoothing directly into the search process, whereas post-hoc smoothing is applied only to the final path after the search is complete.

Due to its efficiency, simplicity and generality, many variants of Theta* have been developed [4], such as Incremental Phi* for unknown 2D grids [6], Theta* on non-uniform costmaps [7], Lazy Theta* in known 3D environments [8] and Strict Theta* that restricts planning to taut paths [9]. A taut path is one whose turns occur only at obstacle corners, tightly wrapping around the obstacles. On a uniform costmap, only such taut paths can be true shortest paths [10]. In addition to Theta*, there are several other any-angle path-planning algorithms, including Field D* [11] and Block A* [12].

However, all of these algorithms still cannot guarantee the discovery of true shortest paths, although they often produce shorter and more realistic paths than traditional A*. To overcome this limitation, Anya [13] has been developed for optimal any-angle path planning. As presented in Fig. 1c, the search nodes of Anya are not located at grid centers or corners, but are instead represented as triangular regions (a *root* point + an *interval* line). The search process resembles shining a flashlight from the start point, reorienting it at each obstacle corner, and continuing until the target point is reached. In this way, Anya limits its search to taut paths, making it both fast and capable of finding true shortest paths on grids.

Anya has been extended from regular grids to irregular navigation meshes through a variant, Polyanya [14]. However, extending Anya to 3D grids or dynamic environments (where dynamic environments refer to those with moving obstacles in this article) remains challenging. While its unique search nodes improve performance, they also hinder its scalability [15]. More analysis and evaluation regarding any-angle path planning can be found in [4,16].

In addition to any-angle path planning on grids, there are other algorithms for the Euclidean Shortest Path Problem (ESPP) that are not based on grids, such as RayScan [17], R2 [18], and End Point Search (EPS) [19]. These algorithms are generally faster than Theta* and Anya, and their performance is independent of grid resolution. However, similar to Anya, they are primarily tailored for solving ESPP in static environments and are not readily adaptable to more complex scenarios involving dynamic obstacles. Given the widespread use of grid representations in the real world and their inherent flexibility in path planning, this research continues to explore grid-based approaches, aiming to develop general techniques for optimal any-angle path planning in both static and dynamic environments.

While many any-angle path-planning algorithms have been proposed, only a few are capable of handling dynamic obstacles, such as Any-Angle Safe Interval Path Planning (AA-SIPP) [20] and Time-Optimal AA-SIPP (TO-AA-SIPP) [21]. AA-SIPP and TO-AA-SIPP can be regarded as any-angle variants of Safe Interval Path Planning (SIPP) [22]. TO-AA-SIPP has also been applied to develop the first optimal any-angle Multi-Agent Path Finding (MAPF) algorithm [23].

In general, optimal any-angle path planning can incur considerable computational overhead for two reasons. First, as movement is not limited to predefined graph edges, a large number of non-adjacent vertices may need to be considered during node expansion, resulting in a large branching factor. Second, establishing such non-local connections requires online visibility checks, and similar visibility computations may be repeatedly performed during the search. To address these issues, we developed an initial version of Zeta*-SIPP in our previous work [24] for dynamic environments by introducing two core techniques: *elliptical forward expansion*, which restricts unnecessary node expansion while preserving optimality, and *field of view*, which reduces redundant computations in visibility checking. This research extends our previous conference paper [24], further illustrates these techniques, introduces additional refinements, and demonstrates their applicability in both static and dynamic environments. As a result, Zeta* and Zeta*-SIPP are developed accordingly.

[1] Screenshots from URL: https://dronectr.tudelft.nl/, ID: pathfinder [2].

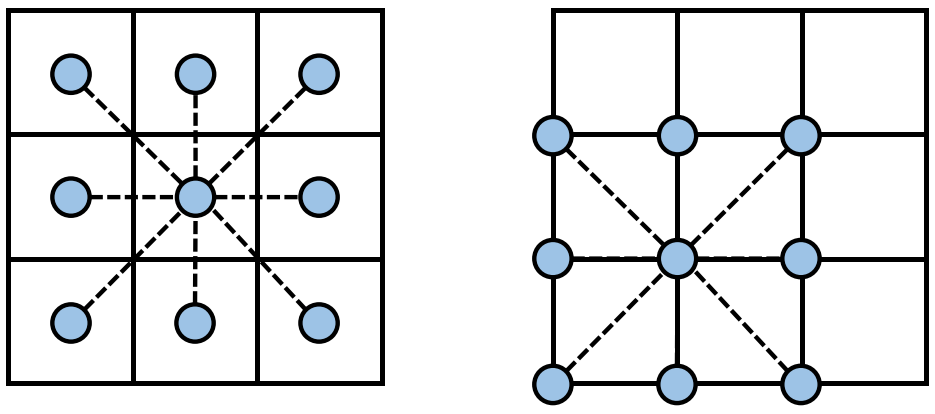

**Fig. 2.** Graph vertices on grids.

Zeta* is tailored specifically for static environments and, like Anya, considers only taut paths to avoid unnecessary node expansions. As a grid-based optimal any-angle path planner, the name Zeta* also pays tribute to Theta*. Overall, our goal for Zeta* is to develop a "classic" grid-based algorithm that retains point-based search nodes while achieving performance comparable to the state-of-the-art planner Anya. This design choice can make Zeta*, in contrast to Anya, more easily extensible to other settings, such as weighted terrains and dynamic environments.

Building on this foundation, we then demonstrate how Zeta* can be extended to Zeta*-SIPP to address environments with dynamic obstacles. Zeta*-SIPP is devised on top of the state-of-the-art optimal planner TO-AA-SIPP, and benchmarking results show that it achieves more than a twenty-fold speedup over TO-AA-SIPP on average. Collectively, Zeta* and Zeta*-SIPP provide a unified and scalable solution for optimal any-angle path planning on grids.

Compared with our previous conference paper [24], the main additional algorithmic contributions of this work can be summarized as follows:

(1) In addition to the inverted scanning method introduced in the previous paper, this research further develops a forward scanning method to achieve field-of-view computation and elliptical forward expansion. Based on these two scanning strategies, we develop two variants of Zeta*-SIPP: Zeta*-SIPP-i and Zeta*-SIPP-f. Zeta*-SIPP-i updates the original (conference) version of Zeta*-SIPP by introducing a bounding box to incrementally expand the *bound* list, while keeping the remaining procedures unchanged. In contrast, Zeta*-SIPP-f is a newly proposed variant that is more memory-efficient and slightly faster than Zeta*-SIPP-i. Additionally, we propose a pruning strategy for inverted expansion based on endpoint occupancy constraints, which can reduce the list of potential parents for open nodes and thus accelerate both TO-AA-SIPP and Zeta*-SIPP.
(2) Elliptical forward expansion and field-of-view computation are not limited to dynamic environments. In this study, we extend these techniques to static environments and integrate them with several pruning strategies, including taut path restriction and obstacle-corner-based node generation, which are applicable only in static settings. The Zeta* algorithm presented in this article is therefore specifically tailored and optimized for static environments and is not obtained by simply removing the SIPP component from Zeta*-SIPP. Based on the two different scanning methods, we introduce two variants of Zeta*: Zeta*-i and Zeta*-f. Zeta*-f is similar to Anya and achieves comparable performance.

Note that, for clarity, this article first presents Zeta* and then Zeta*-SIPP, as the latter is more complex. Accordingly, the article is structured as follows: Section 2 defines path-planning problems in both static and dynamic environments and introduces the relevant concepts and notations. Sections 3 and 4 illustrate the elliptical forward expansion and field of view respectively. Section 5 describes Zeta* for static environments, while Section 6 presents Zeta*-SIPP for dynamic environments. Section 7 discusses the similarities and differences between existing approaches and the proposed algorithms. Section 8 summarizes the findings of this research and outlines several recommendations.

## 2. Preliminaries

Consider an agent navigating from a start point $p_s$ to a target point $p_t$ in a graph $G = (V, E)$ where $V$ denotes the set of vertices and $E$ is the set of edges. On regular grids, the graph vertices can be either the centers or the corners of the grid cells, as illustrated in Fig. 2. Most path-planning algorithms, such as A* and Theta*, are compatible with both configurations. However, certain algorithms, like Anya, rely heavily on taut paths and obstacle corners and are not applicable when the graph vertices are located at grid centers. In dynamic environments, SIPP typically treats a grid cell as a unit and computes the time intervals during which it is occupied by dynamic obstacles. Therefore, to enable meaningful comparisons among different algorithms, this article adopts corner-based vertices for path planning in static environments and center-based vertices for dynamic environments.

On regular grids, the graph edges usually connect adjacent vertices, as shown in Fig. 2. A* searches within this fixed graph structure and thus results in zig-zag path patterns. In contrast, any-angle path planning is not restricted to adjacent connections; instead, it dynamically constructs links between non-adjacent vertices during the search, allowing for straighter, more direct paths. In the extreme worst case, any-angle path planning produces a *fully connected* grid (online), where every pair of grid cells is directly connected unless obstructed by static obstacles. Fortunately, such an extreme worst case is very rare in practice and also depends on the specific algorithm used.

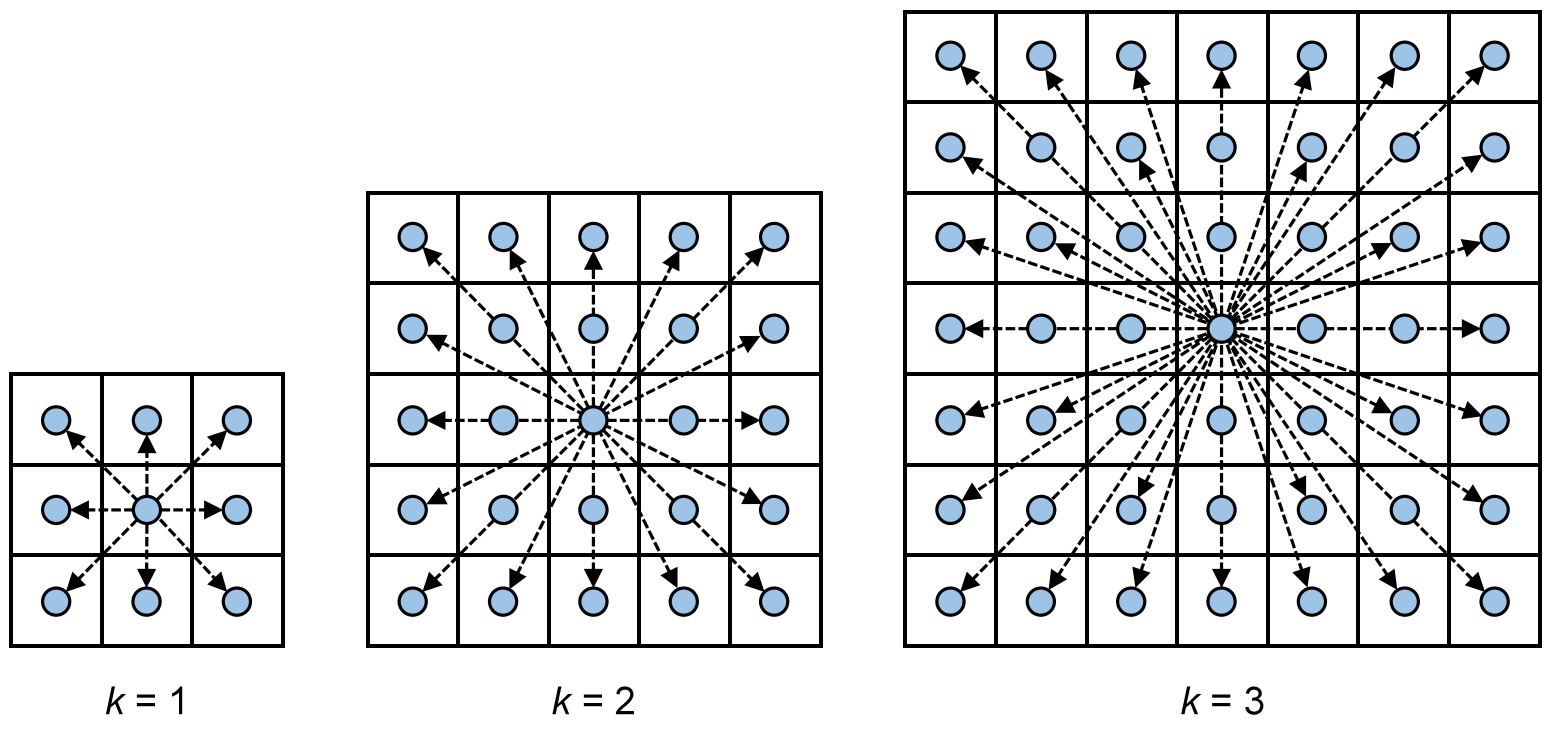

**Fig. 3.** Square neighborhoods of radius $k$ on grids.

The agent is permitted two types of actions during navigation: *move* and *wait*. It can either *wait* at a certain vertex or *move* from one vertex to another. However, in static environments, the *wait* action is meaningless, as it does not help avoid collisions with static obstacles. Therefore, only the *move* action is considered in static scenarios. The agent moves at a constant speed, and the cost of an action (wait or move) is defined by its duration. The agent is allowed to turn or wait only at vertices, and inertial effects are neglected. For simplicity, the agent's radius is ignored; in practice, it can be accounted for by inflating the boundaries of static and dynamic obstacles accordingly.

A *path plan* is an ordered sequence of position-time pairs $\pi = \{(p_1, t_1), (p_2, t_2), \ldots, (p_n, t_n)\}$ where $p_i$ represents a position and $t_i$ denotes the waiting time at the position $p_i$. If the *wait* action is forbidden or unnecessary, then $t_i = 0$ for all $i = 1, 2, \ldots, n$. The *cost* of a path plan is the total duration of all actions, including waiting times. The objective is to find a *time-optimal* plan from a start point $p_s$ to a target point $p_t$. In static environments, the time-optimal path is equivalent to the shortest path, given the assumption of constant agent speed.

In dynamic environments, it is assumed that the plans of dynamic obstacles, denoted by $\{\pi^1, \pi^2, \ldots, \pi^k\}$, are known in advance, and that the obstacles disappear after completing their respective plans. This assumption is reasonable for flying vehicles. For instance, drones typically need to land to reload or recharge after completing their missions. Some studies assume that dynamic obstacles remain at their target locations indefinitely [21], since ground robots cannot simply disappear and will instead become static obstacles upon reaching their destinations. The choice between having dynamic obstacles disappear or become static depends on the specific application and does not affect the core algorithm procedure.

To describe the algorithms proposed in this article, we adopt notation consistent with A*. In A*, a search node $n$ is categorized as either *open* or *closed*. A node is closed when its *best* parent node has been identified. Since only closed nodes can serve as parents of other nodes, a closed node also indicates that the path to it has been finalized. When the target node is closed, the final path has been successfully found. Conversely, an open node means that its parent has not yet been determined and the path to it may still be improved.

Regarding cost functions, $g(n)$ represents the real cost from the start node $n_s$ to the node $n$, and it is initially set to infinity ($\infty$). $g(n) < \infty$ indicates that a feasible path to the node $n$ has been found. $g(n, n')$ denotes the real cost from the node $n$ to another node $n'$, and thus, $g(n) = g(n_s, n)$. The heuristic $h(n)$ refers to the estimated cost from the node $n$ to the target node $n_t$, which is usually set as the duration or length of the direct line between $n$ and $n_t$ in any-angle path planning. $h(n, n')$ is the estimated cost from $n$ to $n'$, and thus, $h(n) = h(n, n_t)$. In A*, the total cost of a node is $f(n) = g(n) + h(n)$.

## 3. Elliptical forward expansion

Optimal any-angle path planning on grids is a non-trivial problem, and only a few algorithms have been developed to address it, with Anya being the most notable example. However, Anya relies heavily on obstacle corners and taut paths, which limits its applicability in dynamic environments. In this section, we revisit basic grid-based path-planning procedures and introduce elliptical forward expansion as a general technique for optimal any-angle path planning.

In grid-based path planning, *neighborhoods* play a crucial role in influencing path quality [25]. A neighborhood primarily refers to the set of nodes an algorithm considers for expansion at each search step. Fig. 3 presents square neighborhoods of radius $k$, which integrate the commonly used 8-connected neighbors. As $k$ increases, the path quality generally improves as well. As $k \to \infty$, all nodes on the grid are considered neighbors of the currently expanded node. A* with $k \to \infty$ is capable of finding optimal any-angle paths on grids.

However, increasing $k$ can also significantly slow down grid-based path planning, as it results in a larger branching factor for search. Although A* with an unbounded neighborhood can achieve optimal any-angle path planning, it may expand many unnecessary nodes that do not contribute to finding the optimal path. For example, nodes located far from the direct line between the start and target points may be redundant to expand, especially in open areas.

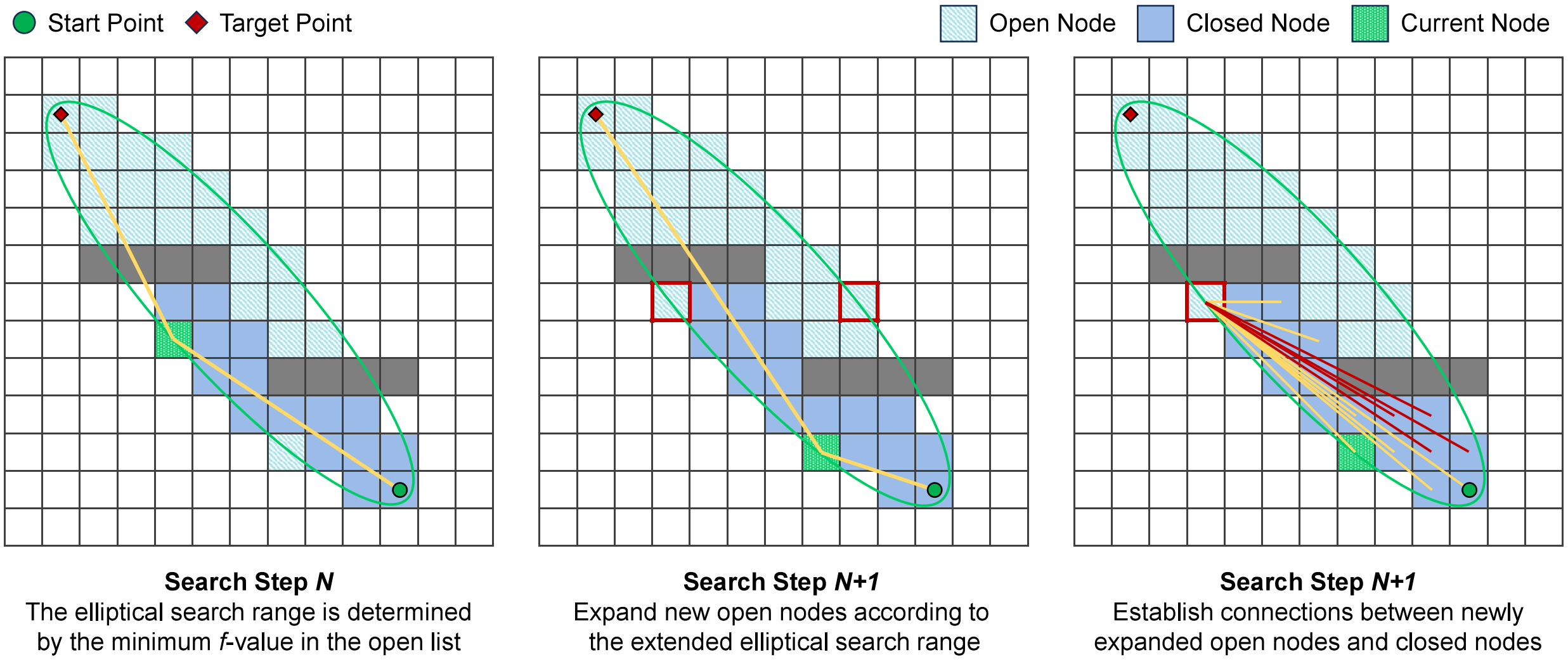

**Fig. 4.** Elliptical forward expansion. The green dot is the start point and the red diamond is the target point. The light blue cells represent open nodes, the dark blue cells indicate closed nodes, and the green cell marks the current node to be inserted into *closed*. A red line indicates an invalid parent-child connection due to obstacles. (For interpretation of the references to color in this figure legend, the reader is referred to the web version of this article.)

To prune such unnecessary nodes while preserving optimality, we introduce a novel expansion strategy to replace traditional square-neighborhood expansion: any-angle forward expansion, or referred to as *elliptical forward expansion*. As indicated in our previous work [24], elliptical forward expansion is inspired by Informed Rapidly-exploring Random Tree* (Informed RRT*) [26] and Batch Informed Trees (BIT*) [27]. It leverages a key geometric property of ellipses: the sum of the distances from any point on the boundary of an ellipse to its two foci is a constant; the sum of the focal distances is smaller for points inside the ellipse and larger for points outside. In path planning, the start and target points can be set as the foci of an ellipse. With its range properly defined, the ellipse can be used as an upper bound for node expansion, with nodes outside this bound (i.e., with larger costs) temporarily excluded from consideration.

Since the search frontier (or range) of A* is usually defined by the minimum $f$-value in the *open* list [28], the major axis length $L$ of the ellipse can also be defined as: $L \geq \min_{n \in open} f(n)$. If $L < \min_{n \in open} f(n)$, then there may exist a node outside the ellipse with an $f$-value lower than the current $\min_{n \in open} f(n)$, violating the required monotonicity for optimal search [28].

For simplicity, we assume $L = \min_{n \in open} f(n)$, and the nodes within the ellipse can be defined by

$$f_{low}(n) \leq L = \min_{m \in open} f(m) \quad \forall n \in S \tag{1}$$

where $f_{low}(n) = h(start, n) + h(n)$, the lower bound of $f(n)$, denotes the length of the major axis (ellipse), and $S$ is the search space. If all nodes satisfying Inequality (1) are inserted into *open*, then

$$\min_{m \in open} f(m) < f_{low}(n) \leq f(n) \quad \forall n \in S_{out} \tag{2}$$

where $S_{out} = S \setminus (open \cup closed)$ is the search space outside the explored region (*open* and *closed*). It indicates that the minimum $f$-value in *open* is also the minimum in the remaining search space ($S \setminus closed$), and thus there is no need to consider $S_{out}$ for optimal search. Please note that the elliptical search range is tied to the minimum $f$-value in *open* and therefore extends whenever this value increases, forming a growing ellipse.

After defining the elliptical search range, the next step is how to expand nodes within this growing ellipse. As demonstrated in Fig. 4, the elliptical boundary is defined by $\min_{n \in open} f(n)$, which is the $f$-value of the current node. As the ellipse extends, new nodes are added to the *open* list, marked by bold red boxes. To assign a parent to each newly added open node, visibility/transition checks are performed. In Fig. 4, a red line indicates an invalid parent-child connection due to obstacles. Note that only closed nodes can serve as parents. Once all potential parents are identified, a comparison is made to select the best parent for each new open node. Unlike square-neighborhood expansion, which is local and centered on the current node, elliptical forward expansion considers the search space from a global perspective.

## 4. Field of view

Any-angle path planning dynamically builds links between non-adjacent vertices during the search process, enabling the discovery of straighter and more direct paths compared to traditional graph-based algorithms. Therefore, online visibility checking is a key factor influencing the performance of any-angle path planning.

In Theta*, a line-of-sight approach is used to check whether the current node's neighbor can be reached directly from the current node's parent. According to the triangle inequality, this direct path is guaranteed to be no longer and often shorter than the one passing through the current node. This approach typically involves applying a line drawing algorithm [29,30] to identify the grid cells traversed by the direct path and to check whether any of them are occupied by obstacles. As the line of sight cannot pass through obstacles, the algorithm terminates upon encountering the first obstructed cell. This line-of-sight check can also be applied to A* with square neighborhoods, as for $k > 1$, connecting non-adjacent neighbors and the current node also requires visibility checks.

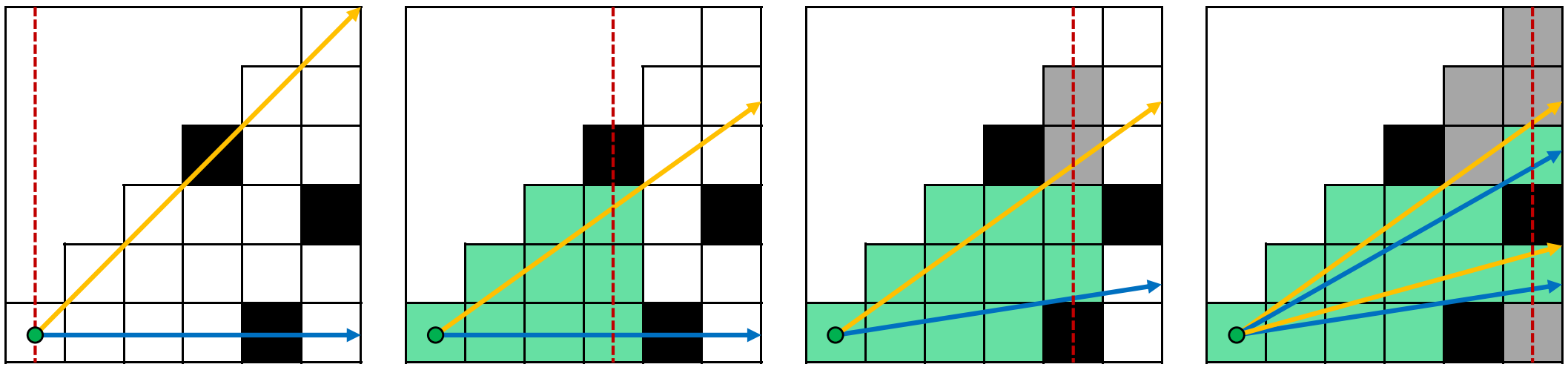

**Fig. 5.** Octant-based symmetric shadowcasting, adapted from [32]. The green dot denotes the origin, and the scan proceeds column by column from left to right. The red dashed line indicates the column currently being scanned. The green cells are visible from the origin while the gray cells are not. (For interpretation of the references to color in this figure legend, the reader is referred to the web version of this article.)

However, the line-of-sight approach may become inefficient when considering large neighborhoods. For instance, in A* with square neighborhoods, if the current node is regarded as a "light source", the surrounding grid cells may be checked multiple times (see the arrows in Fig. 3), leading to redundant scans and wasted resources. To address this issue, we introduce a field-of-view approach and implement *symmetric shadowcasting* [31,32]. Although shadowcasting is a classic algorithm, it remains the state-of-the-art field-of-view method for grids smaller than $512 \times 512$ [33]. Actually, the search node of Anya can also be interpreted as a field of view, which avoids traditional line-of-sight computations and makes Anya fast, especially in open areas.

There are several variants of the shadowcasting algorithm, such as octant-based and quadrant-based shadowcasting. Fig. 5 presents an octant-based example. To construct a field of view, octant-based shadowcasting first divides the area to be scanned into eight octants. Within each octant, the slope range can be normalized to $[0, 1]$ (i.e., $[0°, 45°]$) using a simple linear transformation. In Fig. 5, since nodes are placed at the centers of grid cells, a cell is marked as visible only if its center lies within the *current* slope range. When the scan encounters a blocked cell (black), the slope range is narrowed or split for subsequent columns. This approach reduces repeated grid traversal compared with independent line-of-sight checks from the same "light source".

Shadowcasting is naturally compatible with square neighborhoods in A* due to the way it processes grid-based visibility: radially from a center (current node) outward. For example, when expanding a node with a square neighborhood, shadowcasting with a scan depth of $k$ can be applied to determine the visibility of its candidate neighbors. However, integrating shadowcasting with elliptical forward expansion requires additional effort because the elliptical boundary does not correspond to a fixed radial scan depth on the square grid. As the elliptical frontier extends, the required scan depth varies with direction and must be adjusted accordingly.

Fig. 6 presents two different integration methods: *inverted scanning and forward scanning*. Inverted scanning treats the open node as a "light source". When a new node is added to $open$, shadowcasting is performed to identify nodes visible from it within the elliptical boundary. Due to grid aliasing, a buffer needs to be added around the ellipse to avoid incorrectly classifying cells near the boundary as invisible. This issue arises because the white cells not yet reached by the growing ellipse are treated as obstacles, despite not being true obstacles. For example, in Fig. 6, the buffer can be set to $\sqrt{2}$ grid lengths according to the triangle inequality. In forward scanning, the closed node is treated as a "light source" rather than the open node, and shadowcasting is used to identify nodes visible from it. As the visible nodes are continuously added to the $open$ list with the extension of the elliptical search range, the field of view must also extend to include them. Therefore, an incremental version of shadowcasting is implemented to avoid repeatedly scanning nodes that are already connected.

Fig. 7 illustrates two variants of incremental forward scanning. In naive forward scanning, the bounding box of the elliptical search range is used to limit the scan range. As the ellipse grows, the bounding box is correspondingly pushed outward, thereby progressively increasing the scan depth. However, this approach may process many redundant nodes when the ellipse is rotated. To address this issue, the process can be further constrained by incorporating cost information. In cost-bounded forward scanning, since the $g$-value of a closed node is known, a "local" ellipse can be created by $\min_{m \in open} f(m) - g(n)$. The scan then only needs to extend beyond this local ellipse, rather than the entire "global" ellipse. This approach resembles the search node generation process in Anya (see Fig. 1). However, unlike Anya, shadowcasting does not regard the scan ranges as search nodes, making it more straightforward to understand and implement.

The performance difference between inverted and forward scanning mainly depends on the ratio of open to closed nodes, as this ratio directly affects the number of scans performed. In the extreme worst case where all nodes in the search space are closed, forward scanning could be worse than inverted scanning. This is because incremental forward scanning for each closed node will eventually cover the entire search space, whereas inverted scanning for each node is bounded by the elliptical boundary at each step. However, since forward scanning starts from a closed node whose historical path is determined, information such as the $g$-value and outgoing heading can be exploited to prune the scan range if only static obstacles are present (which will be detailed in the next section).

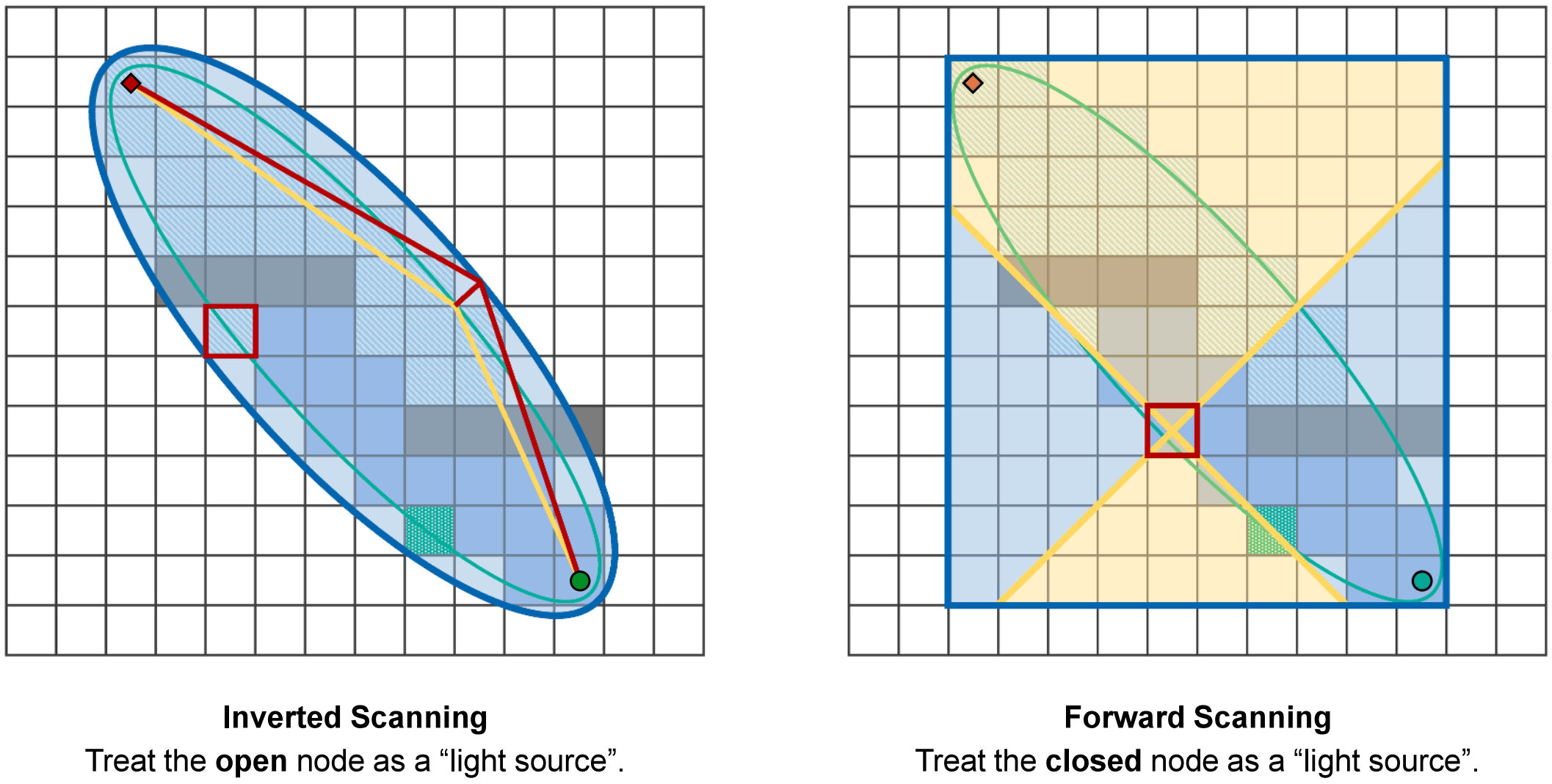

**Fig. 6.** Inverted and forward scanning with elliptical forward expansion. The green dot is the start point and the red diamond is the target point. The gap between the yellow and red lines in the left plot represents the buffer required due to grid aliasing. (For interpretation of the references to color in this figure legend, the reader is referred to the web version of this article.)

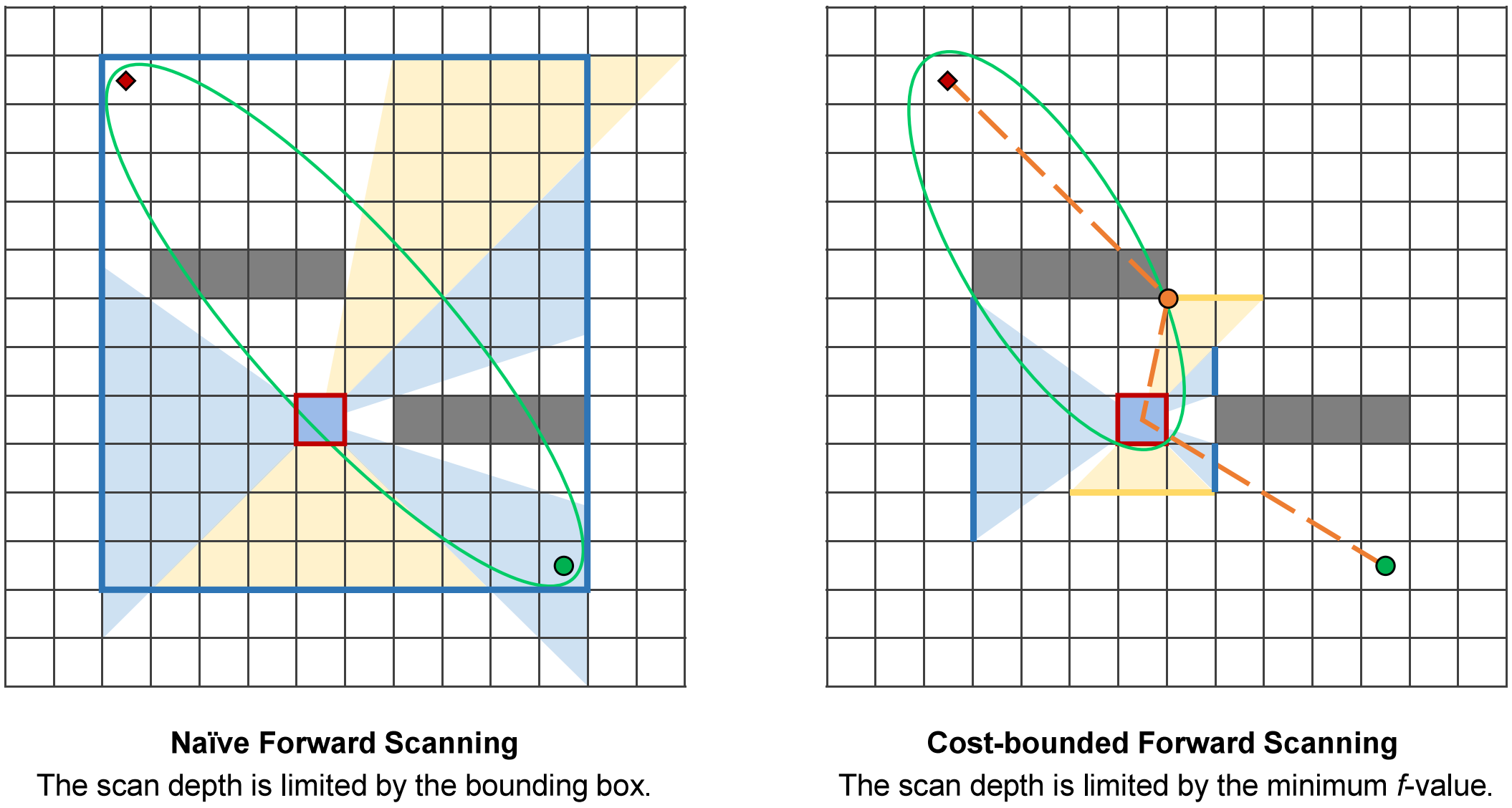

**Fig. 7.** Naive and cost-bounded forward scanning. The blue box in the left plot is the bounding box of the elliptical search range. The blue and yellow triangles correspond to the scan ranges. The foci of the green ellipse in the right plot are the closed node and the target node. (For interpretation of the references to color in this figure legend, the reader is referred to the web version of this article.)

## 5. Zeta* for static environments

By incorporating the elliptical forward expansion and field of view into A*, we introduce Zeta*, a novel optimal any-angle path-planning algorithm for static environments. In the general case, a straightforward implementation of these techniques applies to both center-based and corner-based vertices on grids (recall Fig. 2). To further enhance performance, however, we restrict the search space of Zeta* to only obstacle corners and taut paths, similar to Anya [13] and Strict Theta* [9]. The main loop of Zeta* is shown in Algorithm 1. It closely resembles A*, with the only difference being the forward expansion step. Depending on the scanning method used in this step, we present two variants of Zeta* in this section: Zeta*-i and Zeta*-f.

At a high level, Zeta* consists of three closely connected components. First, it retains the A*-based search backbone (Algorithm 1), in which search nodes are maintained in the *open* and *closed* lists and the node with the minimum $f$-value is selected at each step. Second, elliptical forward expansion determines the current search range according to the minimum $f$-value in *open*. As this value

**Algorithm 1** Zeta* Main Loop.

**while** $open \neq \emptyset$ **do**
    $n \leftarrow \arg\min_{n \in open} f(n)$
    move $n$ from $open$ to $closed$
    **if** $n = target$ **then**
        **return** `pathTo`($n$)
    **end if**
    `forwardExpansion`($n$)
**end while**
**return** $\emptyset$

**Table 1**
Main additional variables introduced in Zeta*.

| Variable | Meaning | Variant |
| --- | --- | --- |
| *bbox* | Axis-aligned bounding box enclosing the growing elliptical search range. | Zeta*-i |
| *bound* | List of nodes within *bbox*, not in *open* or *closed*, awaiting inverted scanning. | Zeta*-i |
| *costBuffer* | Additional cost margin used to reduce repeated updates of *bbox*. | Zeta*-i |
| *costBound* | *costBound* $\leftarrow \min_{n \in open} f(n) +$ *costBuffer*, used to determine the extent of *bbox*. | Zeta*-i |
| $children(n)$ | Nodes visible from $n$ whose connections are recorded during inverted scanning. | Zeta*-i |
| $s$ | A scan range defined by an origin node, scan depth, and slope range. | Zeta*-f |
| *scan* | List of scan ranges maintained for incremental forward scanning. | Zeta*-f |
| $f'(s)$ | Cost of scan range $s$, used to determine its processing order. | Zeta*-f |

increases, the elliptical frontier extends, allowing additional candidate nodes to be considered. Third, field-of-view computation is used to efficiently establish visibility connections between these candidate nodes and closed nodes, enabling the candidate nodes with valid connections to be inserted into or updated in *open* for further search.

The two variants of Zeta*, Zeta*-i and Zeta*-f, share the same A*-based search procedure and elliptical forward expansion principle, but differ in how visibility connections are established. In Zeta*-i, the elliptical frontier is tracked using an axis-aligned bounding box (*bbox*), with a *bound* list maintaining the candidate nodes identified as the bounding box extends. Inverted scanning is then performed from a candidate node to identify visible closed nodes that may serve as its parent. In contrast, Zeta*-f starts from closed nodes. Each closed node maintains incremental scan ranges that are progressively extended as the elliptical frontier extends, and newly visible non-closed nodes are inserted into or updated in *open*. Therefore, the main distinction between Zeta*-i and Zeta*-f is the direction in which visibility connections are established: from candidate nodes toward closed nodes in Zeta*-i, and from closed nodes toward candidate nodes in Zeta*-f.

Zeta* is tailored and optimized for static environments, and the Zeta*-f variant achieves performance comparable to Anya, the current state-of-the-art method. Interestingly, as Zeta* evolves—from inverted scanning to cost-bounded forward scanning—it is increasingly similar to Anya in both design and performance. The introduction of Zeta* thus demonstrates how optimal any-angle path planning can be achieved via elliptical forward expansion and field of view within a traditional grid-based framework. This both deepens the understanding of optimal any-angle path planning and provides a foundation for extending it to other more complex scenarios, such as dynamic environments.

To facilitate algorithmic descriptions, Table 1 summarizes the main additional variables used in Zeta*-i and Zeta*-f. Their detailed roles and update procedures are introduced in Sections 5.1 and 5.2, respectively.

### *5.1. Zeta*-i: Inverted scanning*

To perform inverted scanning, we need to first determine which nodes are expanded by the elliptical search range. The simplest approach may be a brute-force method that compares each node against the minimum $f$-value in *open* using Inequality (1). However, this is inefficient, since it requires computing $f_{low}(n)$ for all nodes and results in many unnecessary comparisons. Therefore, we introduce a new list, called *bound*, to indicate the nodes around the elliptical boundary. The computation of $f_{low}(n)$ and the comparison in Inequality (1) need only be performed within *bound*.

The differences among *bound*, *open*, and *closed* are as follows: When a node $n$ is inserted into *bound*, $f_{low}(n)$ is computed. When the node is moved from *bound* to *open* based on Inequality (1), $f(n)$ is computed. Finally, when it is moved from *open* to *closed*, $f(n)$ is finalized as its true parent is determined. Similar to *open*, which is typically sorted by $f(n)$, *bound* can also be sorted by $f_{low}(n)$. Since the elliptical boundary grows continuously, nodes with lower $f_{low}(n)$ are always inserted into *open* earlier.

To define the *bound* list, we compute the axis-aligned bounding box of the growing ellipse. Let $(x_c, y_c)$ be the center of the ellipse, $a$ the length of the semi-major axis, $b$ the length of the semi-minor axis, and $\theta$ the rotation angle. Then, the parametric form of the ellipse can be given by

$$x(t) = x_c + a\cos t\cos\theta - b\sin t\sin\theta, \quad y(t) = y_c + a\cos t\sin\theta + b\sin t\cos\theta \tag{3}$$

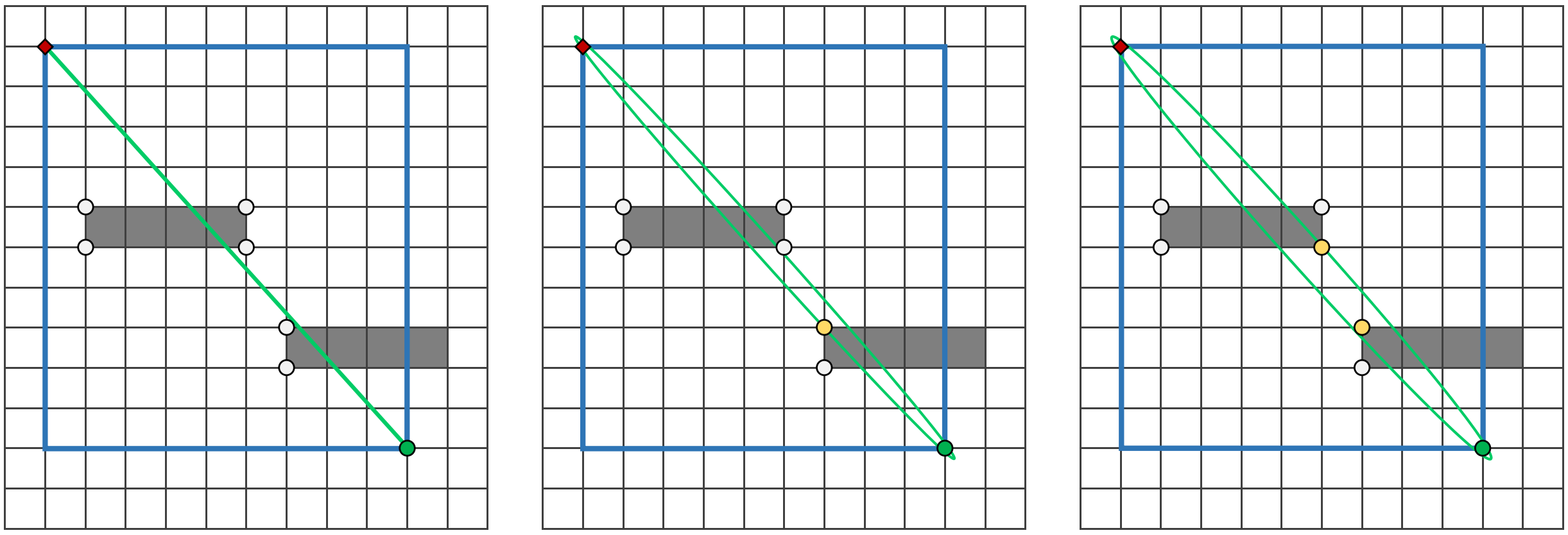

(a) The initial bounding box is defined by the line connecting the start and target. (b) *If the open list is empty*, check the bound node closest to the straight line. (c) If no feasible path exists to the first bound node, proceed to the next one.

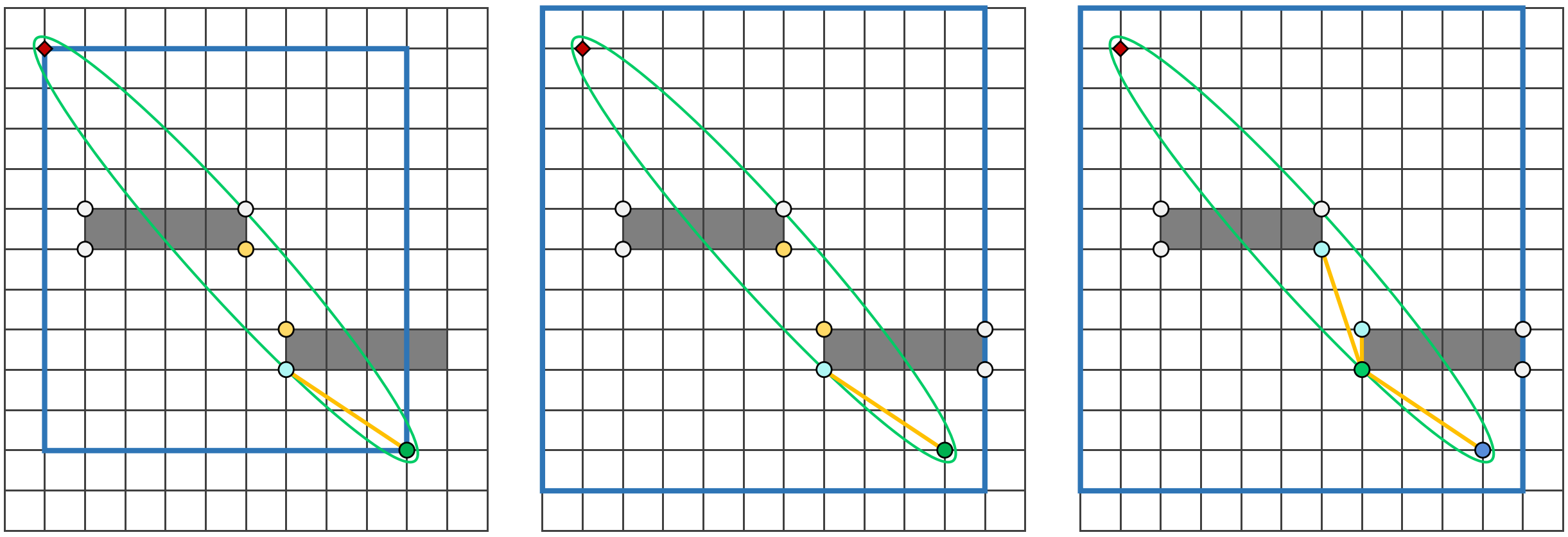

(d) Expand until the open list contains at least one node with a valid *f*-value. (e) Extend the bounding box to cover the new elliptical search range. (f) When a new node is closed, add more open nodes based on stored visibility.

**Fig. 8.** An example illustrating how Zeta*-i works. The blue box represents the bounding box. The green dot denotes the start or current node, while the red diamond denotes the target node. White dots indicate bound nodes. Yellow dots represent nodes that have been removed from the *bound* list, but have not yet been inserted into *open*. Light blue dots correspond to open nodes, whereas dark blue dots correspond to closed nodes. (For interpretation of the references to color in this figure legend, the reader is referred to the web version of this article.)

where $t \in [0, 2\pi)$. Therefore, the half-width $w$ and half-height $h$ of the bounding box can be computed by

$$w = \sqrt{(a\cos\theta)^2 + (b\sin\theta)^2}, \quad h = \sqrt{(a\sin\theta)^2 + (b\cos\theta)^2} \tag{4}$$

This approach allows bound nodes to be easily obtained by slicing the 2D array of nodes, eliminating the need for explicit comparisons using Inequality (1). Since the ellipse grows continuously, we only need to focus on the ring between the old and new bounding boxes to identify and add new bound nodes.

Before diving into the details of the algorithm, we first present an example to illustrate how Zeta*-i works with the bounding box and *bound* list, as depicted in Fig. 8. As shown in Fig. 8a, the initial bounding box is defined by the line segment connecting the start and target nodes during initialization. The nodes within the bounding box are inserted into the *bound* list and marked in white. Note that, in static environments, optimal paths must pass through obstacle corners; therefore, Zeta*-i only generates nodes at obstacle corners, except for the start and target nodes.

Next, as shown in Fig. 8b, if the *open* list is empty, the bound node with the lowest $f_{low}(n)$ should be removed from the *bound* list, and inverted scanning needs to be performed to determine whether connections exist between this node and other closed nodes (at this stage, only the start node is closed). If such a connection exists, this node will be inserted into *open* with its $f$-value computed. The visibility results between this node and other *non-closed* nodes are also recorded for later use. Note that, in addition to the case where the *open* list is empty, this operation should also be performed whenever $f_{low}(n) \leq \min_{m\in open} f(m)$. This process continues until the *open* list contains at least one node with a valid $f$-value (Fig. 8c–d).

Then, as $\min_{n\in open} f(n)$ exists, the bounding box can be updated according to the current elliptical search range, as shown in Fig. 8e. Newly expanded nodes, if any, will be inserted into the *bound* list as well. If there exists a node satisfying $f_{low}(n) \leq \min_{m\in open} f(m)$, that node should be removed from *bound* and inverted scanning needs to be performed. Overall, this operation ensures that no node outside *open* can have an $f$-value lower than $\min_{n\in open} f(n)$.

Therefore, we can proceed to the next step (Fig. 8f). The node with the lowest $f$-value (i.e., the current node) is moved from *open* to *closed*. Since the visibility results between non-closed nodes were recorded during inverted scanning, more nodes can be inserted into *open* based on the current node without performing extra visibility checks. Additionally, when the current node is removed from *open*, $\min_{n\in open} f(n)$ is updated and the bounding box should be extended for further elliptical forward expansion. The overall process continues until a path is found or the entire search space has been explored.

**Algorithm 2** Zeta* with Inverted Scanning.

1: **function** `forwardExpansion`($n$)
2: **for each** $n' \in children(n) \setminus closed$ **do** ▷ $children(n)$ is recorded in the `invertedScan` function.
3: `updateParent`($n, n'$)
4: **end for**
5: **while** $open = \varnothing$ **do**
6: `updateOpenList`($bound$)
7: **if** $bound = \varnothing$ **then**
8: **if** $bbox$ = mapBounds **then**
9: **break**
10: **end if**
11: `extendBoundingBox`($step$) ▷ Extend one step outward since no $\min_{n\in open} f(n)$ is available.
12: `updateBoundList`($bbox$)
13: **end if**
14: **end while**
15: **if** $open \neq \varnothing$ **then**
16: **if** $\min_{n\in open} f(n) > costBound$ **then**
17: $costBound \leftarrow \min_{n\in open} f(n) + costBuffer$ ▷ A buffer can be added to the bounding box.
18: `extendBoundingBoxEllipse`($costBound$)
19: `updateBoundList`($bbox$)
20: **end if**
21: `updateOpenList`($bound$)
22: **end if**
23: **end function**

24: **function** `updateOpenList`($bound$)
25: **while** $bound \neq \varnothing$ and ($open = \varnothing$ or $\min_{n\in bound} f_{low}(n) \leq \min_{n\in open} f(n)$) **do**
26: $n \leftarrow \arg\min_{n\in bound} f_{low}(n)$
27: remove $n$ from $bound$ ▷ $n$ will be added to $open$ if there exists a closed node as its parent.
28: `invertedScan`($n$)
29: **end while**
30: **end function**

31: **function** `invertedScan`($n$)
32: $N_{visible} \leftarrow$ `shadowcasting`($n$) ▷ Be cautious of grid aliasing when performing shadowcasting.
33: **for each** $n' \in N_{visible}$ **do**
34: **if** $n' \in closed$ **then**
35: `updateParent`($n', n$) ▷ If $n$ is not in $open$, it may be inserted within this function.
36: **else**
37: insert $n'$ into $children(n)$ and $n$ into $children(n')$
38: **end if**
39: **end for**
40: **end function**

After illustrating the basic idea of Zeta*-i through an example, the detailed forward expansion procedure is presented in Algorithm 2. Lines 2–4 indicate that when a node is closed, it becomes a potential parent of the nodes visible from it. $children(n)$ stores the visibility results between non-closed nodes and corresponds to Line 37 in the inverted scanning procedure. The `updateParent` function describes a standard parent update procedure with a path tautness check, as shown in Algorithm 3.

Line 5 checks whether the minimum $f$-value exists. If not, the elliptical boundary like Fig. 4 cannot be determined, and the algorithm proceeds with Lines 6–13 repeatedly. The `updateOpenList` function moves nodes from *bound* to *open*, with Line 25 derived from Inequality (1). In Lines 6–13, this function is used only to add more nodes to *open* from the *bound* list. If the *bound* list is empty

**Algorithm 3** `updateParent`$(n, n')$.

> **if** `isTautPath`$(n, n')$ and $g(n) + h(n, n') < g(n')$ **then**
>     $g(n') \leftarrow g(n) + h(n, n')$
>     $parent(n') \leftarrow n$
>     $f(n') \leftarrow g(n') + h(n')$
>     insert or update $n'$ in *open*
> **end if**

(Line 7), the algorithm extends the old bounding box slightly (e.g., by one grid cell) and inserts the newly included nodes into the *bound* list. The `updateOpenList` (Line 6) can then be invoked again if the condition in Line 5 still holds.

Lines 15–22 present the normal case of the elliptical forward expansion procedure. Note that $\min_{n\in open} f(n)$ may decrease during the while-loop within the `updateOpenList` function. This variation does not affect the outcome. As long as the bounding box fully covers the elliptical search range, Inequality (2) is guaranteed to hold. The decrease in $\min_{n\in open} f(n)$ can help prevent unnecessary expansions of the *open* list and reduce the number of visibility checks. Additionally, in practice, a buffer (i.e., *costBuffer*) can be added to the bounding box to reduce the need for repeated executions of `extendBoundingBoxEllipse`, as computing the bounding box also requires time and resources.

The main idea of the `invertedScan` function has been illustrated in Fig. 6. The `shadowcasting` function is bounded by the elliptical search range, determined by the current $\min_{n\in open} f(n)$. Note that only corner nodes with a single blocked adjacent cell will be added to $N_{visible}$ during shadowcasting due to the taut-path constraint. As shown in Lines 33–39, if a visible node $n'$ is closed, it may be the parent of the source node $n$; otherwise, the visible connection is recorded in both $children(n)$ and $children(n')$. These recorded connections are then used in Lines 2–4, eliminating the need to perform extra visibility checking (e.g., shadowcasting) when the current node is closed at each step. The drawback of this operation, however, is that this recording can incur substantial memory overhead on large maps.

### 5.2. Zeta*-f: Forward scanning

Inverted scanning centers on open nodes, aligning with the basic procedure of elliptical forward expansion: when a new node is added to the *open* list, visible connections are then established *from this new open node* to other nodes. However, we can also start from the perspective of closed nodes using forward scanning, ensuring that the scan ranges of all closed nodes cover the newly added open node.

Fig. 9 presents an example illustrating how Zeta*-f works. In this example, quadrant-based shadowcasting is implemented instead of octant-based shadowcasting. As a result, the scan depth in the cost-bounded forward scanning of Zeta*-f is limited only in the upward and downward directions, but not in the left or right directions, making this version more similar to Anya. This can help reduce the number of scan ranges. As shown in Fig. 9a, four scan ranges are initialized for the four quadrants originating from the start node during initialization. The initial elliptical search range can be regarded as the line segment connecting the start and target nodes. The scan range in the second quadrant (upper left) should thus be pushed outward until it lies outside the ellipse. During this process, nodes visible from the start node will be inserted into *open*. Note that once an open node is identified, the elliptical search range may change due to the updated minimum $f$-value in *open*, as shown in Fig. 9b.

Then, we can proceed to the next step (Fig. 9c–e). The node with the lowest $f$-value (i.e., the current node) is moved from *open* to $closed$. Similar to the start node, a scan range is initialized for the current node. However, due to the taut-path restriction, the generated scan range is limited to taut paths only. Since the $g$-value of the current node is fixed, a local elliptical search boundary can be constructed based on the current and target nodes. Every closed node, including the start node, has its own local ellipse. When the current node is removed from *open*, $\min_{n\in open} f(n)$ is updated, and the local ellipse associated with each closed node is extended accordingly. Therefore, the scan ranges from different closed nodes should all be extended beyond their corresponding enlarged local ellipses. During this process, nodes visible from closed nodes will be added to the *open* list. The overall process then repeats from Fig. 9f until a path is found or the entire search space has been explored.

Algorithm 4 introduces the detailed forward expansion procedure of Zeta*-f. The `forwardScan` function pushes the scan ranges beyond the local ellipses, ensuring that the node outside the scan ranges must have a higher $f$-value than $\min_{n\in open} f(n)$. Instead of using a *bound* list, we introduce a $scan$ list to record the scan ranges of incremental shadowcasting. A *scan range* is defined by an origin node, a scan depth, and a slope range (recall Fig. 5). The cost of a scan range can be computed as illustrated in Fig. 10. When a scan range covers a non-closed node, Algorithm 3 is executed to update its parent and/or insert it into *open*.

Readers familiar with Anya may notice that the `forwardScan` function resembles Anya's search procedure. The scan range is similar to Anya's search node $(r, I)$: the origin node corresponds to the root $r$, and the slope range can be converted into the interval $I$. However, in Zeta*-f, we decouple visibility checking from node expansion. This allows us to retain the standard point-based search nodes, as in A* and Theta*, while implementing an Anya-like scanning method to accelerate visibility checks. With shadowcasting and Zeta*-f, the complex notions of Anya's flat and cone nodes, along with its observable and unobservable successors, can be all discarded.

Moreover, the cost functions for scan ranges and search nodes in Zeta*-f do not have to be the same. The cost of a scan range is generally given by the Euclidean distance (or time), whereas the cost of a search node can be modified to incorporate more factors (e.g., waiting time). This flexibility enables Zeta*-f to be readily extended to other settings such as dynamic environments (see Section 6). Essentially, Zeta*-f can be regarded as a generalized version of Anya.

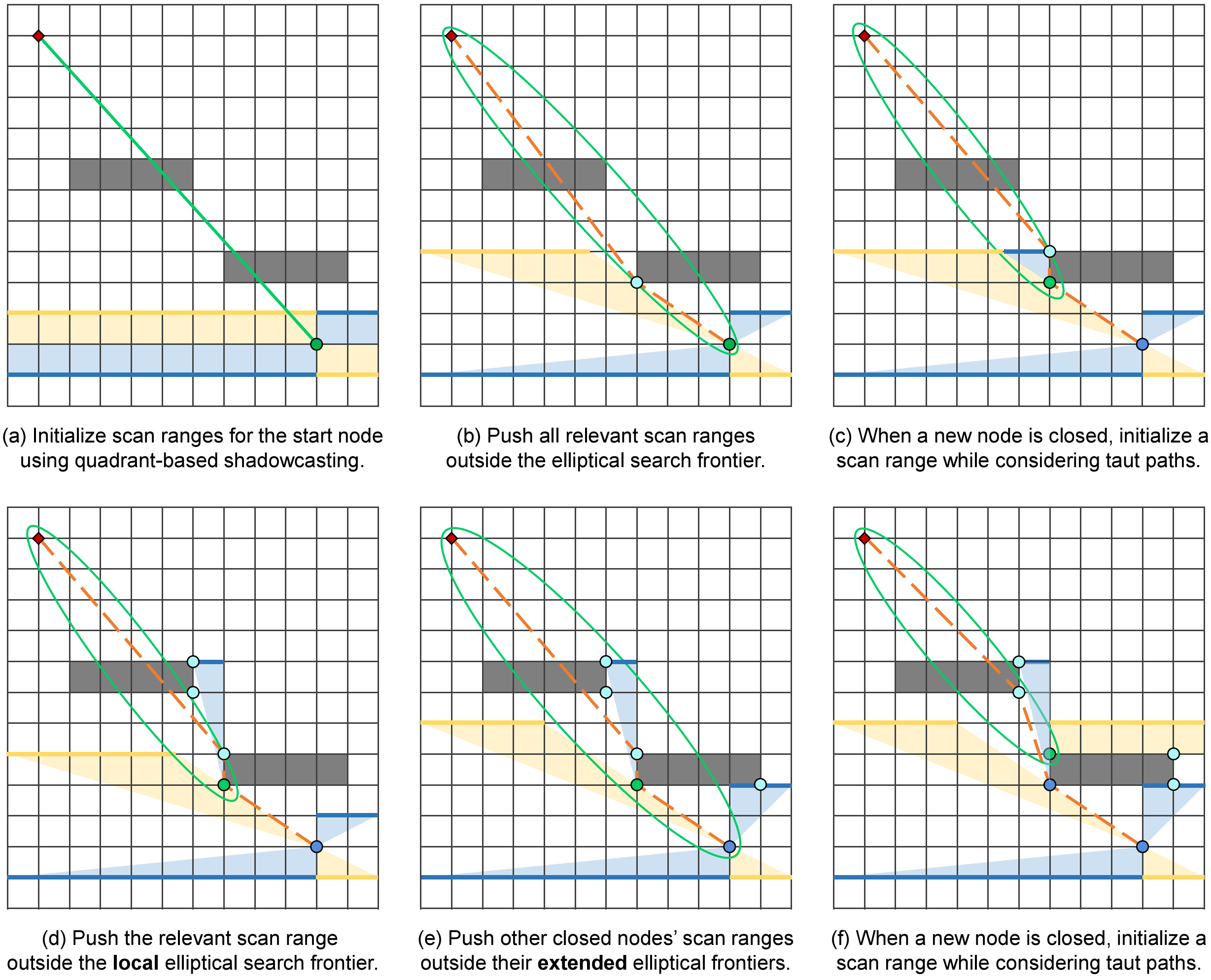

**Fig. 9.** An example illustrating how Zeta*-f works. The green dot denotes the start or current node, while the red diamond denotes the target node. Light blue dots represent open nodes, whereas dark blue dots represent closed nodes. The orange dashed line indicates the current $\min_{n \in open} f(n)$. For other environments (e.g., those with dynamic obstacles), the search nodes are not restricted to obstacle corners, and the ellipse is determined not only by Euclidean distance but also by other factors. (For interpretation of the references to color in this figure legend, the reader is referred to the web version of this article.)

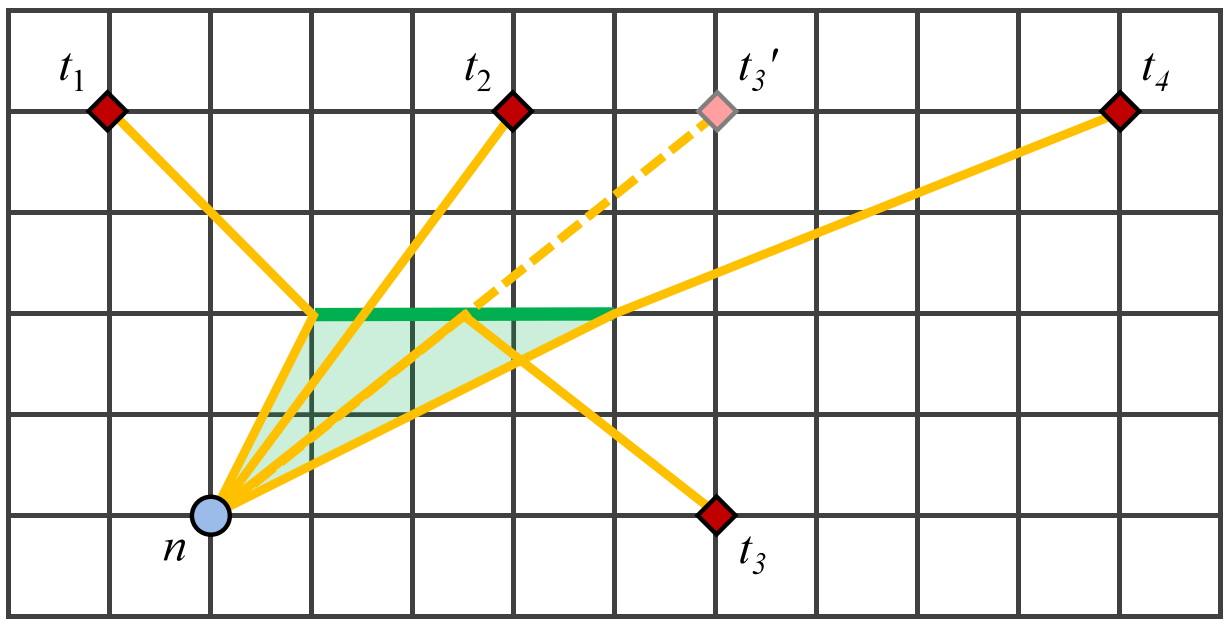

**Fig. 10.** The cost of a scan range $f'(s)$ (green triangle) is computed by $g(n)$ plus the length of the yellow line to the target point $t$. There are four possible relative positions of the target point with respect to the scan range, denoted $t_1$ to $t_4$. $t_3'$ is the mirror position of $t_3$. (For interpretation of the references to color in this figure legend, the reader is referred to the web version of this article.)

**Algorithm 4** Zeta* with Forward Scanning.

**function** `forwardExpansion`($n$)
    $S \leftarrow$ `initScanRange`($n$)    ▷ Initialize the scan range for incremental shadowcasting.
    $scan \leftarrow scan \cup S$
    `forwardScan`($scan$)
**end function**

**function** `forwardScan`($scan$)
    **while** $scan \neq \emptyset$ and ($open = \emptyset$ or $\min_{s \in scan} f'(s) \leq \min_{n \in open} f(n)$) **do**
        $s \leftarrow \arg\min_{s \in scan} f'(s)$
        remove $s$ from $scan$
        $\{S', N_{visible}\} \leftarrow$ `pushScanRange`($s$)    ▷ Push the scan range further by one depth.
        $scan \leftarrow scan \cup S'$
        **for each** $n' \in N_{visible}$ **do**
            **if** $n' \notin closed$ **then**
                `updateParent`($origin(s), n'$)    ▷ The extended scan range covers new non-closed nodes.
            **end if**
        **end for**
    **end while**
**end function**

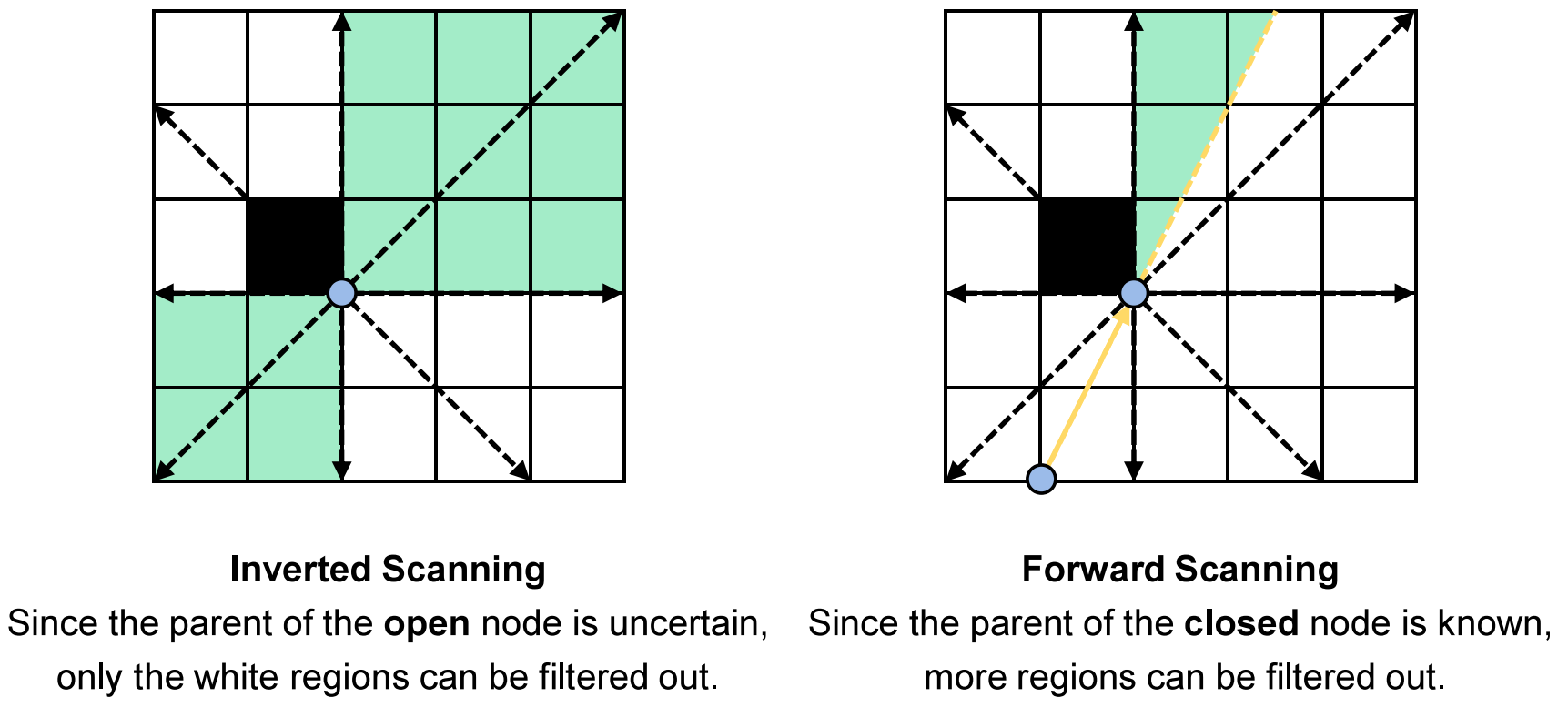

**Fig. 11.** Scan range pruning in inverted and forward scanning.

### 5.3. Scan range pruning

If the path tautness check is ignored and the search considers not only obstacle corners but all possible vertices, Algorithms 2 and 4 can function as general grid-based path-planning algorithms such as A* and Theta*. However, to optimize performance in static environments, the search space is restricted to obstacle corners and taut paths only. In this setting, performance can be further improved by pruning the scan range in shadowcasting. In Zeta*, the scan range can be limited by the heading from the parent node to the node being scanned. As shown in Fig. 11, inverted scanning can explore only the two quadrants along the unblocked diagonal of an obstacle corner. Since the parent of an open node is uncertain, this range cannot be further reduced. In contrast, forward scanning can greatly narrow the scan range using the heading from the parent node because the parent of a closed node has already been determined.

In Zeta*-f, the cost of a scan range will never decrease as its depth increases according to the triangle inequality. For some targets such as $t_2$ in Fig. 10, the cost even remains unchanged. In this case we skip computing the cost and inserting it into the *scan* list, and instead push the scan range further until $t_2$ lies outside the slopes.

### 5.4. Theoretical properties

In this section, we briefly prove the properties of Zeta* (Zeta*-i and Zeta*-f), which are similar to those of A*.

**Lemma 1.** *The node extracted from the open list at each step has the minimum f-value in the search space outside the closed list.*

**Proof.** In Zeta*-i (Algorithm 2), newly included nodes within the bounding box that have not yet been processed are inserted into the *bound* list by the `updateBoundList` function. Then, the `updateOpenList` function ensures

$$\min_{n \in open} f(n) < \min_{n \in bound} f_{low}(n) \leq \min_{n \in bound} f(n) \tag{5}$$

The extendBoundingBoxEllipse function also ensures that every node outside the bounding box $bbox$ has a larger $f$-value than the current minimum $f$-value among open nodes:

$$\min_{n\in open} f(n) < \min_{n\notin bbox} f_{low}(n) \le \min_{n\notin bbox} f(n) \tag{6}$$

Note that $\min_{n\in open} f(n)$ does not increase during expansion. Let $S_{out} = S \setminus (open \cup closed)$ where $S$ is the search space. Let $R$ contain the nodes already removed from $bound$ by inverted scanning but not yet connected to a closed parent (the yellow nodes in Fig. 8); their current $g$-values are $\infty$. Thus $S_{out} \subseteq bound \cup (S \setminus bbox) \cup R$. Based on Inequalities (5) and (6), together with $f(n) = \infty$ for $n \in R$, we have $\min_{n\in open} f(n) < \min_{n\in S_{out}} f(n)$. This concludes the proof for Zeta*-i.

In Zeta*-f (Algorithm 4), the forwardScan function ensures

$$\min_{n\in open} f(n) < \min_{s\in scan} f'(s) \tag{7}$$

Based on the triangle inequality, the cost of a scan range does not decrease as it extends. Therefore, the nodes outside the current scan ranges must have a higher $f$-value than $\min_{s\in scan} f'(s)$. Since the nodes covered by the current scan ranges have already been inserted into $open$ or $closed$, the region outside the scan ranges also indicates $S_{out}$ where $S_{out} = S \setminus (open \cup closed)$. Thus, we have $\min_{n\in open} f(n) < \min_{n\in S_{out}} f(n)$. This concludes the proof. ☐

**Lemma 2.** *When a node is about to be closed, its best parent has been found within the current closed list.*

**Proof.** When a node is about to be closed, it has the minimum $f$-value in the $open$ list (current node). Since only a closed node can be the parent of other nodes, Algorithms 2 and 4 ensure that all visible connections between closed and open nodes are fully established, and Algorithm 3 ensures that the parent of an open node is the best among all closed nodes. If another non-closed node $n'$ were a better parent for the current node $n$, then $g(n') + h(n', n) < g(n)$, which indicates $g(n') + h(n', n) + h(n) < f(n)$. However, according to Lemma 1, we have $f(n) \le f(n') \le g(n') + h(n') \le g(n') + h(n', n) + h(n)$. This contradiction shows that the parent of the current node is the best, not only among all currently closed nodes, but among all nodes in the search space. This concludes the proof. ☐

**Theorem 1.** *Zeta* is complete and optimal.*

**Proof.** Lemma 2 can be interpreted as a recursive formula: Since only a closed node can be the parent of other nodes, when a node is closed, the optimal path to it has been found. Therefore, Zeta* is optimal.

Algorithms 2 and 4 ensure that all visible connections between nodes are established within the elliptical search range (global or local ellipse), which extends until reaching the map boundary. If the target node is reachable, it will eventually be scanned, inserted into $open$, and assigned a parent node. In this case, the main loop will continue until the target node is closed. If the target node is unreachable, the main loop will terminate only when the $open$ is empty. If this condition is met after a node is moved from $open$ to $closed$, then within the forwardExpansion function (e.g., Line 5 in Algorithm 2), the bounding box or scan ranges will be extended in order to add more nodes to $open$. As a result, the main loop terminates when no further nodes can be added. Therefore, Zeta* is complete. ☐

### 5.5. Experiment results

To evaluate the performance of Zeta*-i and Zeta*-f, we compare them with other online grid-based path planners: A* with square neighborhoods ($k = 1, 2, 3$), Theta* and Anya. Polyanya [14], a polygon-based extension of Anya, is also a strong candidate for static environments. However, since our work focuses on grid-based any-angle path planning, it is not included in our evaluation.

All the algorithms are implemented in JavaScript[2] and the experiments were performed on Node.js v22.19.0 on a laptop with 2.30GHz Intel Core i7-11800H and 16 GB RAM, where CPU frequency scaling (Intel Turbo Boost) was enabled. Owing to the single-threaded nature of the Node.js implementation, the processor operated predominantly near its single-core turbo frequency of up to 4.6 GHz. Node.js is a high-performance environment, enabling efficient execution of JavaScript code outside the browser. JavaScript was chosen because its implementations can be seamlessly integrated into web-based pathfinding visualizers, such as PathfindingJS [34], which is beneficial for 'visual' debugging, promotion, and education [2].

In A*-8, the Octile distance is used for the heuristic; Anya applies a heuristic similar to that shown in Fig. 10; for the remaining algorithms, the Euclidean distance is adopted for simplicity. In A*-24 and A*-48, a quadrant-based symmetric shadowcasting, rather than a line-of-sight approach, is implemented for node expansion. Likewise, Zeta*-i and Zeta*-f also employ the quadrant-based shadowcasting for inverted and forward scanning.

The quadrant-based shadowcasting is implemented using a simple linear transformation. During the scanning, all nodes are first mapped into a normalized quadrant with the slope range $[0, \infty]$ and then processed column by column. Although this operation introduces some additional computation, it largely improves the readability and simplicity of the code[2]. Compared with an octant-based approach, the quadrant-based shadowcasting can also reduce edge overlaps and simplify coordinate transformations between real and normalized spaces.

For efficient grid scanning in Anya, Zeta*-i, and Zeta*-f, *bit-packing* is applied. Rather than scanning grid cells individually with a for-loop, groups of cells (e.g., 32 or 64 cells) are represented using a single bit mask, allowing multiple cells to be processed simultaneously through bitwise operations. If a bit mask contains no key events, such as obstacle corners or other critical features, the entire block can be skipped. This approach is particularly effective in static environments, where the waypoints of optimal paths must lie at obstacle corners, enabling large open regions to be traversed efficiently. However, this technique is limited to static scenarios, such as Anya and Zeta*, and does not readily extend to dynamic environments like Zeta*-SIPP.

[2] Our implementation is available at https://github.com/yiyuanzou/zeta-sipp.

**Table 2**
Average runtime of different path-planning algorithms.

| Maps | Average Algorithm Runtime (ms) | | | | | | |
|---|---|---|---|---|---|---|---|
| | A*-8 | A*-24 | A*-48 | Theta* | Anya | Zeta*-i | Zeta*-f |
| BGII | 9.20 | 28.65 | 44.24 | 29.52 | **0.74** | 9.94 | 0.96 |
| DA2 | 5.98 | 14.81 | 21.84 | 11.42 | 1.09 | 8.95 | **1.01** |
| DAO | 12.89 | 30.46 | 45.13 | 27.99 | 4.24 | 27.67 | **4.01** |
| WC3 | 15.95 | 42.79 | 65.31 | 39.29 | **1.36** | 9.35 | 1.44 |
| SC | 49.39 | 141.73 | 231.22 | 152.56 | **5.76** | 63.94 | 6.32 |
| City | 21.16 | 55.98 | 86.19 | 54.55 | 2.11 | 20.18 | **2.09** |
| Maze | 73.41 | 203.43 | 288.51 | 142.58 | **11.66** | 132.06 | 19.08 |
| Rand | **28.27** | 62.99 | 79.54 | 35.57 | 29.93 | 136.82 | 28.86 |
| Room | 37.95 | 109.02 | 169.69 | 64.93 | **6.09** | 19.98 | 7.56 |
| IH | 1851.49 | 6195.09 | 12150.66 | 8625.87 | **54.83** | 2646.96 | 60.98 |

**Table 3**
Average sorted elements: for algorithms other than Zeta*, this metric denotes the number of search nodes.

| Maps | Average Sorted Elements ($\times 10^3$) | | | | | | |
|---|---|---|---|---|---|---|---|
| | A*-8 | A*-24 | A*-48 | Theta* | Anya | Zeta*-i | Zeta*-f |
| BGII | 16.20 | 16.56 | 16.12 | 14.54 | 0.37 | 0.67 | 1.64 |
| DA2 | 6.98 | 7.35 | 7.40 | 6.92 | 0.34 | 0.75 | 1.21 |
| DAO | 15.03 | 15.70 | 15.69 | 14.92 | 1.48 | 1.73 | 4.23 |
| WC3 | 17.72 | 18.96 | 18.50 | 16.80 | 0.31 | 0.64 | 1.71 |
| SC | 60.05 | 62.41 | 61.28 | 58.70 | 2.23 | 4.59 | 7.28 |
| City | 21.03 | 22.77 | 21.58 | 18.96 | 0.68 | 1.75 | 2.21 |
| Maze | 131.28 | 132.05 | 132.07 | 131.62 | 7.23 | 23.86 | 28.77 |
| Rand | 32.00 | 34.14 | 32.73 | 31.40 | 21.45 | 45.79 | 32.39 |
| Room | 40.07 | 43.89 | 43.09 | 40.81 | 2.01 | 4.04 | 8.05 |
| IH | 1130.41 | 1217.83 | 1151.64 | 1104.79 | 14.31 | 28.31 | 69.08 |

In Anya, both cul-de-sac pruning and intermediate pruning, as described by Harabor et al. [13], are implemented to reduce the size of its $open$ list. In Zeta*-i, the *costBuffer* for the bounding box is set to 10 grid lengths based on small-scale tests. The buffer for inverted scanning to address grid aliasing is set to $2\sqrt{2}$ grid lengths, since the nodes are located at grid corners rather than at grid centers. While we refer to $open$, $bound$ and $scan$ as lists in this article, they are actually implemented as *binary heaps* rather than plain lists for sorting operations.

To conduct the experiments, we selected 9 benchmark sets from the Moving AI Lab [35]: Baldurs Gate II (**BGII**), Dragon Age 2 (**DA2**), Dragon Age: Origins (**DAO**), Warcraft III (**WC3**), Starcraft (**SC**), **City** maps, **Maze** maps, **Random** maps (with 10%, 20%, 30% and 40% randomly blocked grid cells) and **Room** maps. All maps, except for DA2, DAO and SC, are of size $512 \times 512$. The first 5 benchmark sets correspond to game maps. Moreover, we included the Iron Harvest (**IH**) game dataset [36], which contains larger maps, with the largest measuring $4760 \times 6400$ and $5728 \times 6336$. In total, the datasets used in the experiments comprise approximately 1.54 million start-target pairs.

Tables 2–4 show the benchmarking results of different algorithms in terms of runtime, number of sorted elements, and number of scanned vertices. For algorithms other than Zeta*-i and Zeta*-f, the sorted elements refer to search nodes. In Zeta*-i, an additional list, denoted as $bound$, is introduced and requires sorting similar to the $open$ list. In Zeta*-f, a $scan$ list is introduced, which also needs to be sorted. Since the search nodes are located at the vertices of grid cells, we record the number of scanned vertices to reflect the visibility checking workload. For the line-of-sight operation in Theta*, this metric is approximated by counting the scanned grid cells.

For A* with square neighborhoods, it can be clearly observed that the runtime increases as $k$ increases. However, the number of sorted elements (i.e., search nodes) does not increase significantly and even decreases in some cases. The runtime slowdown with larger $k$ values is therefore primarily caused by intensive visibility checking rather than by sorting.

Theta* runs at a speed comparable to A*-24, even though it scans a substantially larger number of vertices. This is because line-of-sight checking processes each grid more quickly than shadowcasting. In line-of-sight checking, the line slope remains fixed, whereas in shadowcasting, the slope range has to be recalculated whenever a new blocked cell is encountered. Furthermore, the shadowcasting we implement needs to perform coordinate transformations for each node due to the quadrant normalization, which introduces additional computational overhead. Please note that the implementation of A* with square neighborhoods presented in this article is merely a simple and direct extension of the standard A* algorithm on regular grids, combined with shadowcasting. More advanced optimization techniques can be found in the work of Rivera et al. [37].

**Table 4**
Average scanned vertices: for Theta*, this metric is approximated by the number of scanned grid cells.

| Maps | Average Scanned Vertices ($\times 10^5$) | | | | | | |
|---|---|---|---|---|---|---|---|
| | A*-8 | A*-24 | A*-48 | Theta* | Anya | Zeta*-i | Zeta*-f |
| BGII | 1.21 | 3.52 | 6.47 | 37.34 | 0.32 | 4.13 | 0.27 |
| DA2 | 0.50 | 1.47 | 2.74 | 7.80 | 0.11 | 1.41 | 0.10 |
| DAO | 1.10 | 3.25 | 6.15 | 23.43 | 0.52 | 5.11 | 0.41 |
| WC3 | 1.33 | 4.05 | 7.47 | 41.82 | 0.41 | 4.74 | 0.34 |
| SC | 4.55 | 13.75 | 26.12 | 185.64 | 1.53 | 44.29 | 1.31 |
| City | 1.58 | 4.88 | 8.74 | 67.85 | 0.60 | 16.00 | 0.47 |
| Maze | 8.39 | 21.84 | 38.33 | 95.54 | 1.10 | 4.90 | 1.68 |
| Rand | 1.71 | 4.13 | 5.78 | 4.60 | 0.70 | 5.92 | 0.79 |
| Room | 2.90 | 8.86 | 16.20 | 39.41 | 0.76 | 2.79 | 0.71 |
| IH | 89.10 | 285.35 | 534.31 | 14033.72 | 43.26 | 3658.98 | 33.60 |

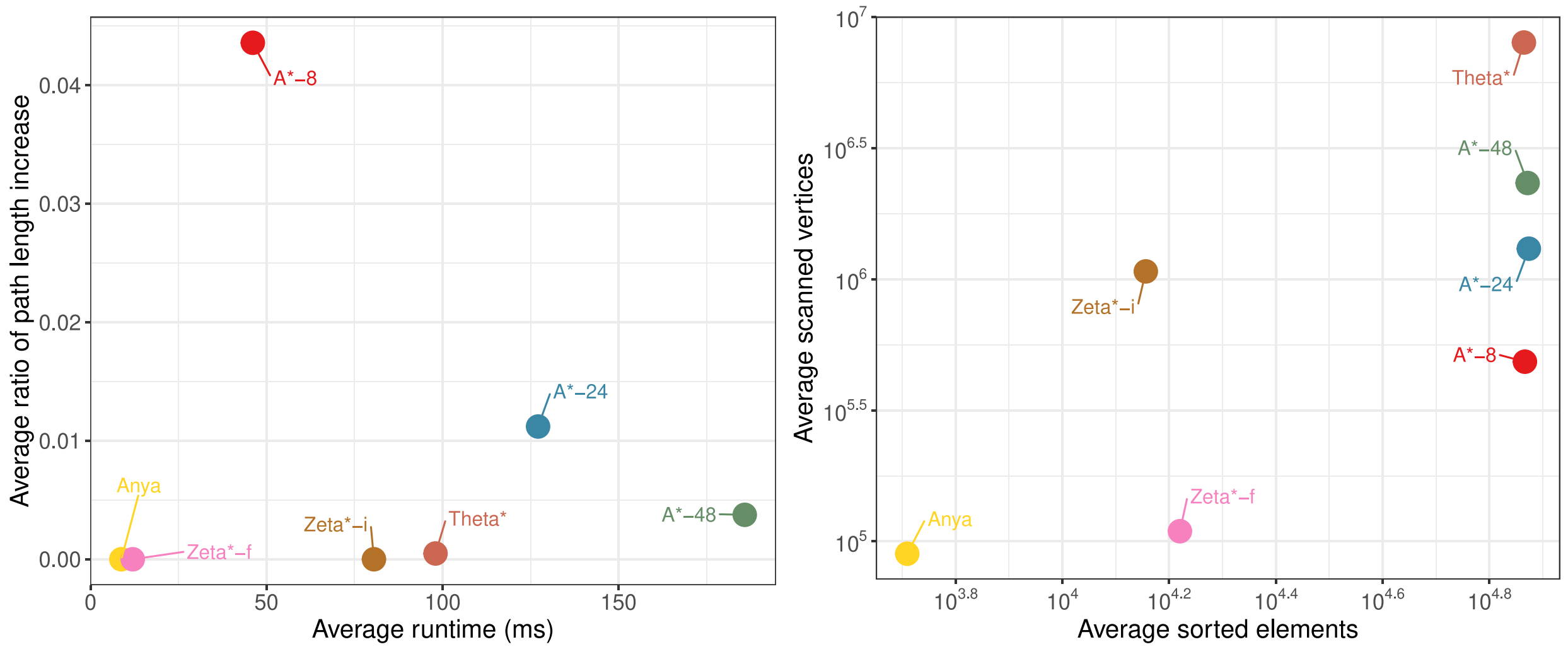

**Fig. 12.** Scatter plots of path length vs. runtime and scanned vertices vs. sorted elements. The path length increase is given as a ratio relative to the true shortest path. Both axes in the right plot are shown on a logarithmic scale. IH results are excluded from this figure because IH maps are significantly larger than the other maps, most of which are $512 \times 512$.

Anya outperforms the other algorithms in most maps. This can be attributed to its reduced search space derived from taut paths and its field-of-view-like search nodes. Zeta*-f closely resembles Anya, especially when implemented with quadrant-based shadowcasting for cost-bounded forward scanning. Zeta*-f separates visibility checking from node expansion, which is beneficial for general grid-based path planning. However, this design requires maintaining two sorted lists instead of one, making Zeta*-f slightly slower than Anya—but still significantly faster than Theta*.

Zeta*-i is much slower than Zeta*-f. This slowdown mainly results from the differences in their scan range pruning (see Fig. 11), which cause Zeta*-i to scan substantially more vertices than Zeta*-f. Therefore, Zeta*-i may not be well suited for any-angle path planning in static environments when considering only taut paths.

Fig. 12 presents the overall benchmarking results of different path-planning algorithms. In general, algorithms that require fewer scanned vertices and sorted elements tend to run faster. Compared with A*-8, A*-24 achieves a notable improvement in path quality. Theta* generates paths that are only slightly longer than the true shortest paths, while running generally faster than A*-24. Among the optimal planners, Anya remains the best performer, while the proposed Zeta*-f also demonstrates comparable performance.

## 6. Zeta*-SIPP for dynamic environments

The proposed elliptical forward expansion and field of view can also be extended to dynamic environments [24]. SIPP is one of the most popular and efficient methods for handling dynamic obstacles. Compared to Anya, Zeta* is easier to integrate with SIPP due to its point-based search nodes and decoupled visibility checking. Accordingly, Zeta*-SIPP is developed for optimal any-angle path planning in the presence of dynamic obstacles. In this setting, the search space can no longer be restricted to taut paths and obstacle corners, and the search nodes are placed at the centers of grid cells. Since the *wait* action is allowed, the goal of Zeta*-SIPP is not to find the shortest path, but the time-optimal conflict-free path.

### 6.1. Safe intervals

The main difference between static and dynamic obstacles on grids lies in their representations. Static obstacles can be directly represented as blocked cells, while dynamic obstacles require consideration of the time dimension. A common approach to handling the time dimension is to divide it into equal-length time slots, constructing a space-time grid map [38]. A search algorithm such as A* can then be applied to this map to find optimal conflict-free paths. However, this approach aggravates the *curse of dimensionality* problem, as it introduces an additional dimension to the search space. To address this issue, Safe Interval Path Planning (SIPP) [22] has been developed, which merges consecutive obstacle-free time slots into *safe intervals*, creating a *compact* space-time map to narrow the search space.

In any-angle path planning, safe-interval-based conflict detection for the direct path between two nodes is similar to line-of-sight checks with static obstacles. First, a line drawing algorithm [29,30] is applied to identify the grid cells traversed by the direct path and computes the time intervals during which each cell is occupied. Then, these intervals are compared against the corresponding *safe intervals* at each cell to determine whether a conflict exists. If a conflict is detected, it can be resolved by either waiting at the parent node for a certain duration or selecting an alternative path (a different parent node) with a lower cost. Therefore, the key to Zeta*-SIPP is to integrate Safe-Interval-based Conflict Detection and Resolution (SI-CDR) into Zeta*.

### 6.2. Inverted expansion

However, unlike line-of-sight checks, SI-CDR is generally more time-consuming, especially when the *wait* action is allowed. There is also no field-of-view approach available for dynamic obstacles. Therefore, integrating SI-CDR directly into the node expansion process of any-angle path planning may significantly slow down the algorithm. To mitigate this issue, *inverted expansion* was introduced in Time-Optimal Any-Angle SIPP (TO-AA-SIPP) [21], which reduces the number of SI-CDR executions by delaying them until necessary.

Inverted expansion is the counterpart of forward expansion: while forward expansion identifies the potential child nodes of a closed node, inverted expansion stores the potential parent nodes of an open node. For example, in A*, when a node is moved from $open$ to $closed$, its eight adjacent neighbors are expanded (i.e., forward expansion). This new closed node thus becomes a *potential parent* of its neighbors, and a comparison like Algorithm 3 is conducted to determine whether this potential parent is better than the current parent of each expanded neighbor. However, instead of determining the neighbor's parent immediately, inverted expansion stores all of its potential parents and defers the comparison. This allows the real cost between a node and its potential parent to be calculated at a later stage.

Let $pp(n)$ denote the potential parent of a node $n$ and $g(pp(n), n)$ the real cost from $pp(n)$ to $n$. Then $g(pp(n)) + g(pp(n), n)$ is the real cost from the start node to $n$ via $pp(n)$. Among all potential parents of $n$, the one with the lowest cost from the start node to $n$ is the *best potential parent* $bpp(n)$. The real cost $g(n)$ can thus be written as

$$g(n) = g(bpp(n)) + g(bpp(n), n) \tag{8}$$

However, in inverted expansion, the real cost $g(pp(n), n)$ is not computed immediately when $n$ is expanded by $pp(n)$. Instead, the heuristic $h(pp(n), n)$, given by the Euclidean distance divided by speed, is used to estimate this cost and serves as its lower bound. Therefore, the cost function is reformulated as

$$f(n) = g_{low}(n) + h(n) = g(bpp(n)) + h(bpp(n), n) + h(n) \tag{9}$$

where $g_{low}(n)$ is the lower bound of $g(n)$ and $h(n)$ is the estimated cost from $n$ to the target node. The best potential parent $bpp(n)$ is then the parent that yields the lowest $g_{low}(n)$ for $n$. Based on this new $f(n)$, once the node with the minimum $f$-value is removed from $open$, instead of directly inserting it into $closed$, SI-CDR is executed to compute the real cost $g(bpp(n), n)$ and determine whether this node can truly be closed (whether $bpp(n)$ is the true parent of $n$). If not, this node has to be re-inserted into $open$ with a new best potential parent.

In this way, SI-CDR is performed only for the most promising node (i.e., the node with the minimum $f$-value), thereby reducing the total number of executions and conserving computing resources. This "lazy" operation can also be applied to visibility checks with static obstacles, like Lazy Theta* [8]. However, in Zeta*-SIPP, it is used only for SI-CDR with dynamic obstacles. There are two reasons: First, visibility checks can be implemented efficiently using line-of-sight or field-of-view approaches on 2D grids. Second, the "lazy" operation introduces additional sorting overhead in $open$, resulting in a trade-off between conflict detection and sorting computation.

### 6.3. Inverted expansion pruning

For SIPP-based planners, a node $n$ can be represented by $(p, [t_1, t_2])$ where $p$ is the location of $n$ and $[t_1, t_2]$ is the corresponding safe interval. This implies that the path from a potential parent node $pp(n)$ to $n$ is valid only if the agent arrives at $n$ from $pp(n)$ within the interval $[t_1, t_2]$. Therefore, $g_{low}(n)$ can be replaced by

$$g'_{low}(n) = \max\{t_1, g_{low}(n)\} = \max\{t_1, g(bpp(n)) + h(bpp(n), n)\} \tag{10}$$

in computing $f(n)$ (Eq. (9)). Moreover, if a potential parent $pp(n)$ has $g(pp(n)) + h(pp(n), n) > t_2$, then the agent from $pp(n)$ can never reach $n$. This connection can thus be pruned to reduce the list of potential parents for $n$.

**Algorithm 5** Zeta*-SIPP Main Loop.

```
while open ≠ ∅ and min_{n∈open} f(n) < ∞ do
    n ← findNextClosedNode(open)                    ▷ This function may return null
    if n = target then
        return pathTo(n)
    end if
    nodeExpansion(n)                                ▷ Integrate both forward and inverted expansion.
end while
return ∅
```

In the original TO-AA-SIPP, Eq. (10) is applied only during initialization. Here, we extend this pruning to the entire search process and further strengthen it by incorporating the occupancy constraints of both endpoint cells. Specifically, Eq. (10) accounts only for the time at which the agent reaches the center of the destination cell. In practice, however, the agent enters the destination cell before reaching its center and continues to occupy the source cell for some time after leaving its center. These endpoint occupancy constraints enable additional pruning of infeasible connections. We denote the resulting arrival-time bound by endpointLowerBound($n'$, $n$), which returns $\infty$ if the endpoint constraints make the connection infeasible or if the resulting lower bound exceeds the verified arrival time $g(n)$. In either case, $n'$ cannot serve as a potential parent of $n$ and can therefore be pruned.

We apply this enhanced pruning to all algorithms that use inverted expansion in this study, including TO-AA-SIPP and Zeta*-SIPP. Consequently, our benchmarking compares Zeta*-SIPP against an improved version of TO-AA-SIPP that incorporates this enhanced pruning, rather than against the original TO-AA-SIPP.

### 6.4. *Zeta*-SIPP-i and Zeta*-SIPP-f*

By integrating SIPP and inverted expansion into Zeta*, we develop Zeta*-SIPP. Similar to Zeta*-i and Zeta*-f, using two different scanning methods, we derive Zeta*-SIPP-i and Zeta*-SIPP-f.

The pseudocode of Zeta*-SIPP is shown in Algorithms 5–10. In the main loop, when a node $n$ is removed from $open$, it is not immediately placed into $closed$. Instead, the findNextClosedNode function (Algorithm 8) is implemented to determine whether $n$ can be closed. If $n$ needs to be re-inserted into $open$, this function will return $null$. Compared with Zeta*, Zeta*-SIPP applies the nodeExpansion function (Algorithms 6 and 7) to integrate both forward and inverted expansion. It is nearly identical to the forwardExpansion function in Algorithms 2 and 4, except that updateParent is replaced by addPotentialParent (Algorithm 10). Note that Zeta*-SIPP-i and Zeta*-SIPP-f differ only in the nodeExpansion function; all other steps are identical. The scan range pruning proposed for Zeta* (Fig. 11) is not applicable to Zeta*-SIPP.

Algorithm 8 is adapted from TO-AA-SIPP [21], with the differences marked in red. The $potentialParents(n)$ refers to the list of potential parents, as defined in Algorithm 10. Line 4 indicates that the best potential parent $bpp(n)$ should be removed from $potentialParents(n)$ to avoid repeatedly assigning the same potential parent as $bpp(n)$ in Algorithm 9. The validateTransition function implements SI-CDR, where the real cost $g_{new}$ is calculated as $g_{low}(n) = g(bpp(n)) + h(bpp(n), n)$ plus the waiting time at $bpp(n)$ required to resolve conflicts with dynamic obstacles. Therefore, $g_{new} \geq g_{low}(n)$. Lines 7–10 update the parent of $n$ if $bpp(n)$ is better.

The newBestPotentialParentExists function checks whether there is a new *best potential parent* $bpp(n)$ that is likely to be better for $n$ than its current parent $parent(n)$, as presented in Algorithm 9. In Line 11 of Algorithm 8, if a new $bpp(n)$ exists, the current node $n$ will be re-inserted into $open$ with this updated $bpp(n)$. In addition to the check on $bpp(n)$, a new condition $g(n) = \infty$ is added to check whether $n$ has a parent. $g(n) = \infty$ indicates $g_{new} = \infty$, meaning the waiting time is infinite and thus the conflicts cannot be resolved. Note that the order of the conditions in Line 11 cannot be changed, as the first condition updates $bpp(n)$ for the next iteration.

In Line 15 of Algorithm 8, the conditions are used to check whether the current node remains the most promising candidate to be the next closed node. For Zeta*-SIPP with inverted scanning, the $minCost$ is

$$minCost = \min\{\min_{n\in open} f(n), \min_{n\in bound} f_{low}(n)\} \tag{11}$$

where $f_{low}(n)$ is the lower bound of $f(n)$. For Zeta*-SIPP with forward scanning, the $minCost$ is

$$minCost = \min\{\min_{n\in open} f(n), \min_{s\in scan} f'(s)\} \tag{12}$$

where $f'(s)$ is the cost of a scan range (recall Fig. 10). If $g(n) + h(n) \leq minCost$, then we have

$$g(n) + h(n) \leq \min_{m\notin closed} f(m) \leq \min_{m\notin closed}(g(m) + h(m)) \tag{13}$$

It ensures that no other non-closed node has a smaller value of $g(n) + h(n)$. If the condition holds, the current node can be closed; otherwise, it has to be re-inserted into $open$. A violation of the condition indicates either that the algorithm needs to expand the search to include more nodes in $open$, or that the node *currently* with the minimum $f$-value may have a lower $g(n) + h(n)$. Note that $f(n)$ in Zeta*-SIPP is different from that in Zeta*, as shown in Eq. (9), and thus $g(n) + h(n)$ is not represented by $f(n)$.

Zeta*-SIPP is built on SIPP and thus its search nodes differ from those in Zeta*. For the same location or grid cell $p$, multiple nodes may exist with different safe intervals. This may lead to repeated visibility checks between the same pair of grids but different pairs of nodes. To mitigate this issue, it is recommended to perform visibility checks and other spatial computations at the grid level rather than the node level whenever possible.

**Algorithm 6** Zeta*-SIPP with inverted scanning.

```
function nodeExpansion(n)
    if n ≠ null then
        for each n' ∈ children(n) \ closed do          ▷ children(n) is recorded in the invertedScan function.
            addPotentialParent(n, n')
        end for
    end if
    while open = ∅ or min_{n∈open} f(n) = ∞ do
        updateOpenList(bound)
        if bound = ∅ then
            if bbox = mapBounds then
                break
            end if
            extendBoundingBox(step)          ▷ Extend one step outward since no min_{n∈open} f(n) is available.
            updateBoundList(bbox)
        end if
    end while
    if open ≠ ∅ and min_{n∈open} f(n) < ∞ then
        if min_{n∈open} f(n) > costBound then
            costBound ← min_{n∈open} f(n) + costBuffer          ▷ A buffer can be added to the bounding box.
            extendBoundingBoxEllipse(costBound)
            updateBoundList(bbox)
        end if
        updateOpenList(bound)
    end if
end function

function updateOpenList(bound)
    while bound ≠ ∅ and (open = ∅ or min_{n∈bound} f_low(n) ≤ min_{n∈open} f(n)) do
        n ← arg min_{n∈bound} f_low(n)
        remove n from bound          ▷ n will be added to open if there exists a closed node as its potential parent.
        invertedScan(n)
    end while
end function

function invertedScan(n)
    N_visible ← shadowcasting(n)          ▷ Be cautious of grid aliasing when performing shadowcasting.
    for each n' ∈ N_visible do
        if n' ∈ closed then
            addPotentialParent(n', n)          ▷ If n is not in open, it may be inserted within this function.
        else
            insert n' into children(n) and n into children(n')
        end if
    end for
end function
```

In our previous work [24], we also developed TO-AA-FoV-SIPP by applying the field-of-view technique to TO-AA-SIPP without elliptical forward expansion. Similar to Zeta*-SIPP, TO-AA-FoV-SIPP uses field-of-view scanning to reduce redundant visibility computations. The main difference is that Zeta*-SIPP further restricts the search and scan ranges through elliptical forward expansion, whereas TO-AA-FoV-SIPP considers all nodes in the search space as potential neighbors for node expansion. When the elliptical search range extends to cover most of the relevant search space, the exploration results of Zeta*-SIPP may approach those of TO-AA-FoV-SIPP. Their performance differences are further evaluated in Section 6.6.

### 6.5. Theoretical properties

In this section, we briefly prove the properties of Zeta*-SIPP, which are similar to those of TO-AA-SIPP.

**Lemma 3.** *When a node is about to be closed, it has the minimum $g + h$ in the search space outside the closed list.*

**Proof.** For Zeta*, Lemma 3 is equivalent to Lemma 1, since in Zeta* the node with the minimum $f$-value is inserted into *closed* immediately after being removed from *open*, and $f(n) = g(n) + h(n)$.

**Algorithm 7** Zeta*-SIPP with Forward Scanning.

**function** `nodeExpansion`($n$)
  **if** $n \neq null$ **then**
    $S \leftarrow$ `initScanRange`($n$) ▷ Initialize the scan range for incremental shadowcasting.
    $scan \leftarrow scan \cup S$
  **end if**
  `forwardScan`($scan$)
**end function**

**function** `forwardScan`($scan$)
  **while** $scan \neq \emptyset$ and ($open = \emptyset$ or $\min_{s \in scan} f'(s) \leq \min_{n \in open} f(n)$) **do**
    $s \leftarrow \arg\min_{s \in scan} f'(s)$
    remove $s$ from $scan$
    $\{S', N_{visible}\} \leftarrow$ `pushScanRange`($s$) ▷ Push the scan range further by one depth.
    $scan \leftarrow scan \cup S'$
    **for each** $n' \in N_{visible}$ **do**
      **if** $n' \notin closed$ **then**
        `addPotentialParent`($origin(s), n'$) ▷ The extended scan range covers new non-closed nodes.
      **end if**
    **end for**
  **end while**
**end function**

**Algorithm 8** `findNextClosedNode`($open$).

1: $n \leftarrow \arg\min_{n \in open} f(n)$
2: remove $n$ from $open$
3: **if** $bpp(n) \in potentialParents(n)$ **then**
4:   remove $bpp(n)$ from $potentialParents(n)$ ▷ Avoid reassigning the same potential parent as $bpp(n)$
5: **end if**
6: $g_{new} \leftarrow$ `validateTransition`($bpp(n), n$) ▷ Compute the real cost based on safe intervals.
7: **if** $g_{new} < g(n)$ **then**
8:   $g(n) \leftarrow g_{new}$
9:   $parent(n) \leftarrow bpp(n)$
10: **end if**
11: **if** `newBestPotentialParentExists`($n$) or $g(n) = \infty$ **then** ▷ The current parent is not the most promising.
12:   insert $n$ into $open$
13:   **return** $null$
14: **end if**
15: **if** $g(n) + h(n) \leq minCost$ **then** ▷ $minCost$ is obtained by Eq. (11) or (12) according to the scanning method.
16:   insert $n$ into $closed$
17:   **return** $n$
18: **else**
19:   insert $n$ into $open$
20:   **return** $null$
21: **end if**

However, for Zeta*-SIPP, $f(n) = g_{low}(n) + h(n)$. When a node is removed from $open$, SI-CDR is performed to compute the real cost $g(n)$, which may be larger than $g_{low}(n)$. As presented in Inequality (13), Line 15 in Algorithm 8 ensures that $g(n) + h(n) \leq \min_{m \in S \setminus closed} f(m) \leq \min_{m \in S' \setminus closed}(g(m) + h(m))$. This concludes the proof. □

**Lemma 4.** *When a node is about to be closed, its best parent has been found given the current closed list.*

**Proof.** Lemma 4 is identical to Lemma 2, which is the key to proving the optimality of A*-based algorithms. Algorithm 10 keeps the best potential parent of each open node updated and Algorithm 9 ensures that the current node's parent remains its best potential parent after SI-CDR. Therefore, when the algorithm reaches Line 15 in Algorithm 8, the current node's parent is guaranteed to be the best among all closed nodes. Lemma 3 then ensures that no other non-closed node can be a better parent for the current node. This concludes the proof. □

**Theorem 2.** *Zeta*-SIPP is complete and optimal.*

**Algorithm 9** `newBestPotentialParentExists`($n$).

$g_{low}(n) \leftarrow g(n)$, $bpp(n) \leftarrow parent(n)$
$f(n) \leftarrow g_{low}(n) + h(n)$
$isExisting \leftarrow false$
**for each** $n' \in potentialParents(n)$ **do**
  $g_{low} \leftarrow$ `endpointLowerBound`$(n', n)$
  **if** $g_{low} < g_{low}(n)$ **then**
    $g_{low}(n) \leftarrow g_{low}$
    $bpp(n) \leftarrow n'$
    $f(n) \leftarrow g_{low}(n) + h(n)$
    $isExisting \leftarrow true$
  **end if**
**end for**
**return** $isExisting$

**Algorithm 10** `addPotentialParent`($n, n'$).

$g_{low} \leftarrow$ `endpointLowerBound`$(n, n')$ ▷ Endpoint pruning for inverted expansion; return $\infty$ if infeasible.
**if** $g_{low} < \infty$ **then**
  insert $n$ into $potentialParents(n')$
  **if** $g_{low} < g_{low}(n')$ **then**
    $g_{low}(n') \leftarrow g_{low}$
    $bpp(n') \leftarrow n$
    $f(n') \leftarrow g_{low}(n') + h(n')$
    insert or update $n'$ in $open$
  **end if**
**end if**

**Proof.** The proof is analogous to that of Zeta*, as the elliptical forward expansion and visibility checking procedures are unchanged. The optimality of Zeta*-SIPP is a direct corollary of Lemma 4. As all potential parents of a node are stored, Zeta*-SIPP will, in the worst case, examine every possibility to determine whether this node is reachable in the presence of dynamic obstacles. If a node is unreachable from all closed nodes, no feasible path exists. Otherwise, as the closed list grows, the algorithm will eventually find a feasible path. Therefore, Zeta*-SIPP is complete. □

### 6.6. Experiment results

To evaluate the performance of Zeta*-SIPP-i and Zeta*-SIPP-f, we compare them with other SIPP-based planners: SIPP with square neighborhoods ($k = 1, 2, 3$), AA-SIPP, TO-AA-SIPP, TO-AA-FoV-SIPP. All the algorithms are implemented in JavaScript[2] and the experiments were performed on Node.js v22.19.0 on a laptop with 2.30GHz Intel Core i7-11800H and 16 GB RAM, the same as in the previous section.

SIPP with square neighborhoods directly extends SIPP by integrating square neighborhoods of radius $k$. While no prior research has explored this variant, it provides a straightforward method to extend SIPP to any-angle path planning. For simplicity, we do not apply shadowcasting or inverted expansion to this variant. AA-SIPP can be regarded as a combination of Theta* and SIPP, which straightens paths by checking whether the current node's neighbors can be reached from the current node's parent with a lower cost. TO-AA-SIPP is the only optimal planner aside from the algorithms proposed in this article. It can be seen as SIPP with an unbounded neighborhood ($k \to \infty$) combined with inverted expansion. TO-AA-FoV-SIPP is a direct extension of TO-AA-SIPP that incorporates the Field of View (FoV) and can be readily implemented on top of it. TO-AA-FoV-SIPP has been shown to be efficient as well in our previous work [24]. Zeta*-SIPP integrates Zeta* and SIPP, as well as inverted expansion. Zeta*-SIPP-i and Zeta*-SIPP-f are two variants of Zeta*-SIPP.

The agent speed is set to 0.1 grid lengths per second to prevent it from skipping cells. For SIPP-8, we use the Octile distance divided by the speed as the heuristic, while for the other variants, we use the Euclidean distance divided by the speed. For TO-AA-FoV-SIPP and Zeta*-SIPP, we implement a quadrant-based symmetric shadowcasting similar to Zeta*. The cost buffer for Zeta*-SIPP-i is the same as that of Zeta*-i (10 grid lengths), but divided by the speed, while the scan buffer is set to $\sqrt{2}$ grid lengths because the nodes are placed at grid centers rather than grid corners.

To conduct the experiment, we chose 6 benchmark maps from the Moving AI Lab [39]: 1) **Empty-48-48**, a $48 \times 48$ map without any static obstacles; 2) **Random-64-64-10**, a $64 \times 64$ map with 10% randomly blocked cells; 3) **Room-64-64-16**, a $64 \times 64$ map divided into 16 rooms; 4) **Maze-128-128-10**, a $128 \times 128$ maze map with 10-wide corridors; 5) **Warehouse-10-20-10-2-2**, a $170 \times 84$ map from the logistics domain; 6) **Berlin_1_256**, a $256 \times 256$ real-world city map. These maps were chosen to represent a variety of environments, use cases, and scales.

**Table 5**
Average sorted elements: for algorithms other than Zeta*-SIPP, this metric denotes the number of search nodes.

| Maps | Average Sorted Elements ($\times 10^3$) | | | | | | | |
|---|---|---|---|---|---|---|---|---|
| | SIPP-8 | SIPP-24 | SIPP-48 | AA-SIPP | TO-AA-SIPP | TO-AA-FoV-SIPP | Zeta*-SIPP-i | Zeta*-SIPP-f |
| Empty | 0.24 | 0.35 | 0.43 | 0.22 | 2.75 | 2.75 | 1.17 | 1.64 |
| Rand | 0.32 | 0.46 | 0.53 | 0.32 | 2.29 | 2.23 | 1.87 | 2.62 |
| Room | 1.56 | 1.70 | 1.79 | 1.57 | 2.60 | 2.58 | 4.61 | 47.26 |
| Maze | 6.77 | 7.32 | 7.49 | 6.96 | 9.20 | 9.23 | 19.53 | 303.23 |
| Ware | 1.00 | 1.36 | 1.43 | 1.14 | 5.37 | 5.39 | 5.62 | 13.63 |
| Berlin | 4.31 | 4.94 | 4.94 | 4.11 | 20.61 | 20.73 | 24.63 | 179.95 |

**Table 6**
Average scanned grids for different path-planning algorithms.

| Maps | Average Scanned Grids ($\times 10^4$) | | | | | | | |
|---|---|---|---|---|---|---|---|---|
| | SIPP-8 | SIPP-24 | SIPP-48 | AA-SIPP | TO-AA-SIPP | TO-AA-FoV-SIPP | Zeta*-SIPP-i | Zeta*-SIPP-f |
| Empty | 0.08 | 0.59 | 1.23 | 0.41 | 373.79 | 12.58 | 7.44 | 3.81 |
| Rand | 0.11 | 0.81 | 1.63 | 0.45 | 390.78 | 3.51 | 3.24 | 1.25 |
| Room | 0.88 | 5.83 | 13.81 | 4.90 | 3883.62 | 32.46 | 27.34 | 32.56 |
| Maze | 4.10 | 23.59 | 56.36 | 35.04 | 93435.17 | 341.07 | 280.72 | 336.00 |
| Ware | 0.38 | 2.43 | 4.98 | 3.12 | 5654.45 | 41.20 | 20.10 | 33.29 |
| Berlin | 2.56 | 15.30 | 34.87 | 42.71 | 434771.90 | 1512.86 | 725.29 | 493.87 |

For each map, 20,000 test cases were generated as follows: 1) Choose 25 benchmark scenario sets (random). 2) For each scenario set, take the last 200 scenarios as test cases and the first 32, 64, 96, or 128 scenarios as dynamic obstacles. 3) Generate the trajectories of dynamic obstacles sequentially using Zeta*-SIPP-f, which are time-optimal, conflict-free and contain any-angle moves. As mentioned in Section 2, we ignore the agent's radius for simplicity. To compensate for this, we extend the time intervals during which dynamic obstacles occupy each grid cell by 10 units. This buffer corresponds to $1/\text{speed}$; since the speed is $0.1$, the result is $10$.

Figs. 13 and 14 show the average runtime of the tested algorithms under different dynamic obstacle conditions. Unlike path planning in static environments, there is no taut-path constraint to restrict the search space. As a result, optimal planners are typically much slower than suboptimal ones. For clarity, the results are presented in two separate figures. As shown, the runtime of all algorithms increases with the number of dynamic obstacles, with SIPP-48 and the optimal planners exhibiting the most significant growth. This is because, to avoid dynamic obstacles, SIPP-48 and the optimal planners generally need to expand more nodes to find (sub-)optimal paths.

To facilitate the analysis of runtime performance, Tables 5 and 6 show the numbers of sorted elements and scanned grids. Since the search nodes of the SIPP-based planners are placed at grid centers, scanned grids are reported instead of scanned vertices. Each value in the tables represents the mean over all dynamic obstacle settings. Interestingly, the number of search nodes in TO-AA-SIPP and TO-AA-FoV-SIPP is not exactly the same, although TO-AA-FoV-SIPP only modifies the visibility checking method. This is because the time-optimal any-angle path is not unique. When waiting time is taken into account, multiple time-optimal path plans can coexist in the solution space. The visibility checking method influences the order in which nodes are examined and thus also the path-planning results. There may be a way to prune these identical paths, but this lies beyond the scope of this article.

As shown in Fig. 13, SIPP-24 performs considerably slower than SIPP-8. This slowdown occurs because SIPP-24 scans many more grids than SIPP-8, owing to its larger neighborhood size. In terms of overall algorithm speed, SIPP-24 performs comparably to AA-SIPP and achieves better performance on larger-scale maps. This demonstrates the potential of SIPP-24 for any-angle path planning in dynamic environments.

As shown in Fig. 14, TO-AA-SIPP is significantly slower than the other optimal planners. This is mainly because it examines visibility connections from each closed node to all other nodes in the search space, and implements line-of-sight rather than field-of-view scanning to perform this process. TO-AA-FoV-SIPP mitigates the latter issue and greatly reduces the number of scanned grids, while Zeta*-SIPP addresses both issues by extending TO-AA-FoV-SIPP with elliptical forward expansion.

However, this does not imply that TO-AA-FoV-SIPP is always slower than Zeta*-SIPP. In the Room and Maze maps, TO-AA-FoV-SIPP outperforms both Zeta*-SIPP-i and Zeta*-SIPP-f. This is because these maps are composed of narrow exits and passages, where the elliptical search range of Zeta*-SIPP often extends to touch the walls. In such cases, Zeta*-SIPP generates a similar number of nodes as TO-AA-FoV-SIPP, and the Zeta*-SIPP-f variant becomes equivalent to it. However, the additional computational overhead introduced by the elliptical forward expansion makes Zeta*-SIPP slower. In most cases, especially in open areas, the elliptical search range typically occupies only a small portion of the overall search space, allowing Zeta*-SIPP to outperform TO-AA-FoV-SIPP in practice.

Zeta*-SIPP-f is clearly faster than Zeta*-SIPP-i in the Empty, Random, and Berlin maps, while showing comparable performance in the other maps. Although Zeta*-SIPP-f may sort far more elements and scan more grids than Zeta*-SIPP-i, their overall runtimes are still similar. This is likely due to the extension of the bounding box in Zeta*-SIPP-i, which also incurs computational overhead. Note that Zeta*-SIPP-i, like Zeta*-i (Line 37 in Algorithm 2), must record visibility connections between non-closed nodes to avoid repeated scanning; otherwise, its runtime would be much slower. When extended to larger maps, Zeta*-SIPP-i may encounter memory limitations, while Zeta*-SIPP-f remains more practical. From a memory perspective, Zeta*-SIPP-f is therefore also more efficient than Zeta*-SIPP-i.

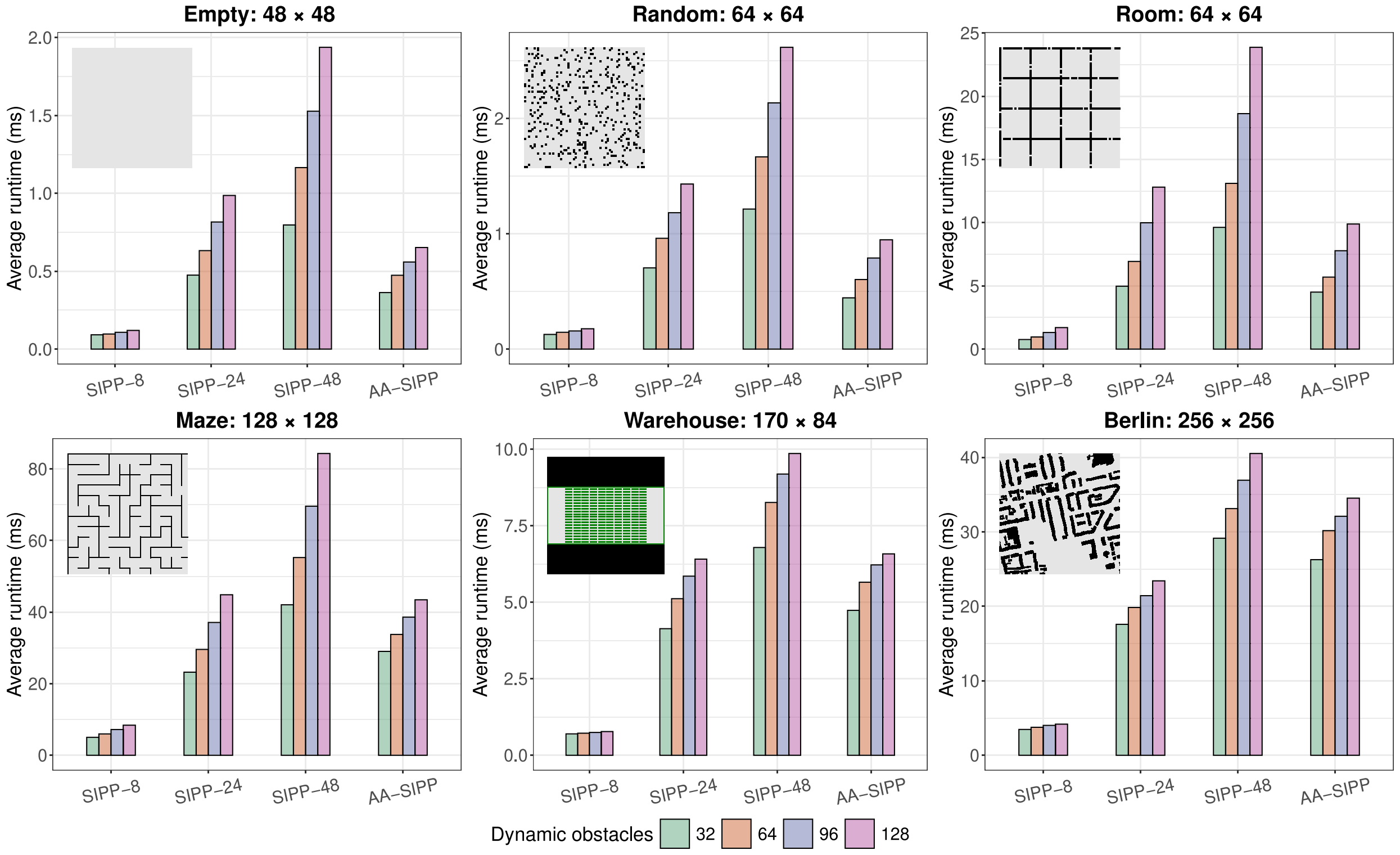

**Fig. 13.** Average runtime of the **suboptimal** algorithms under different dynamic obstacle conditions.

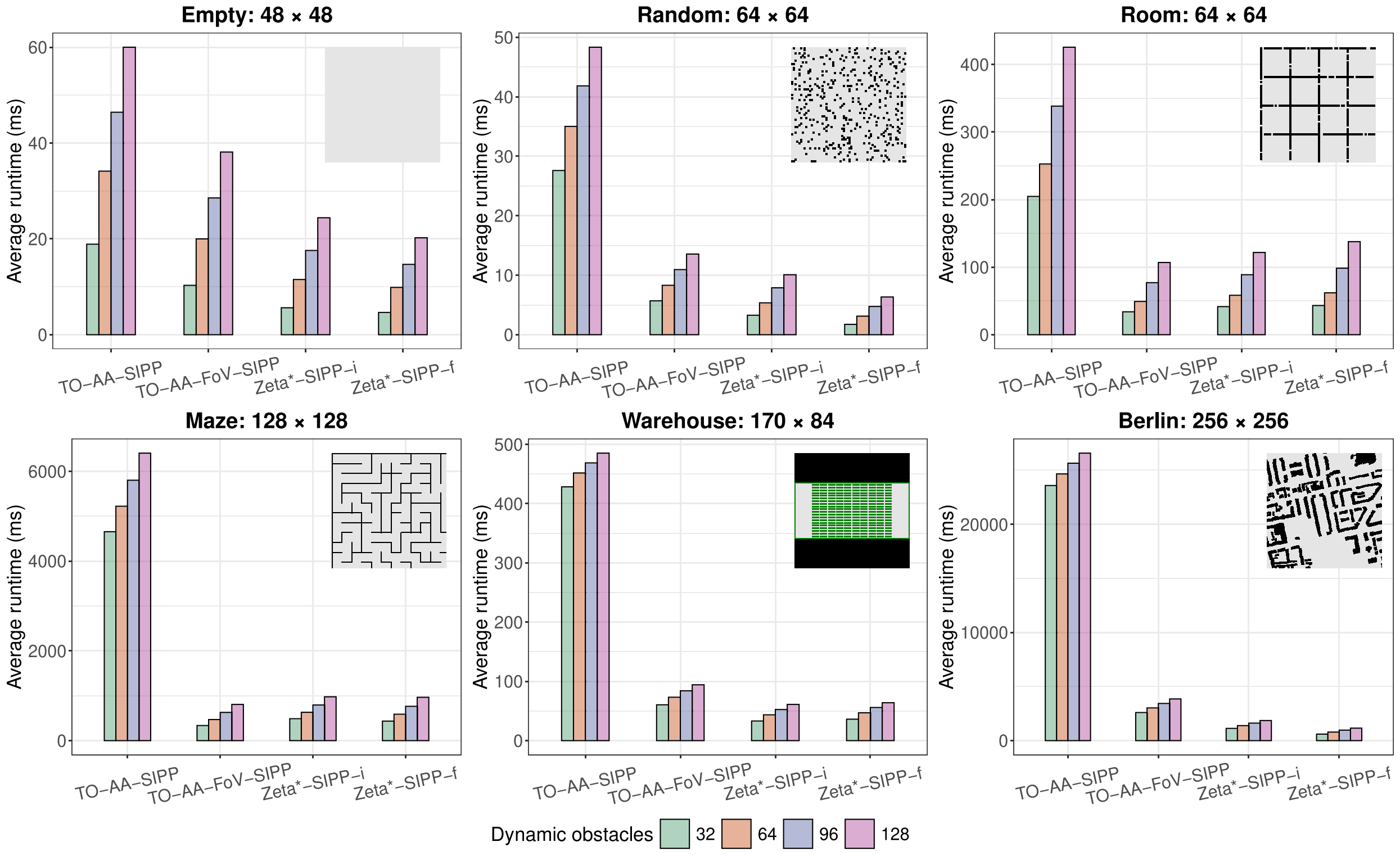

**Fig. 14.** Average runtime of the **optimal** algorithms under different dynamic obstacle conditions.

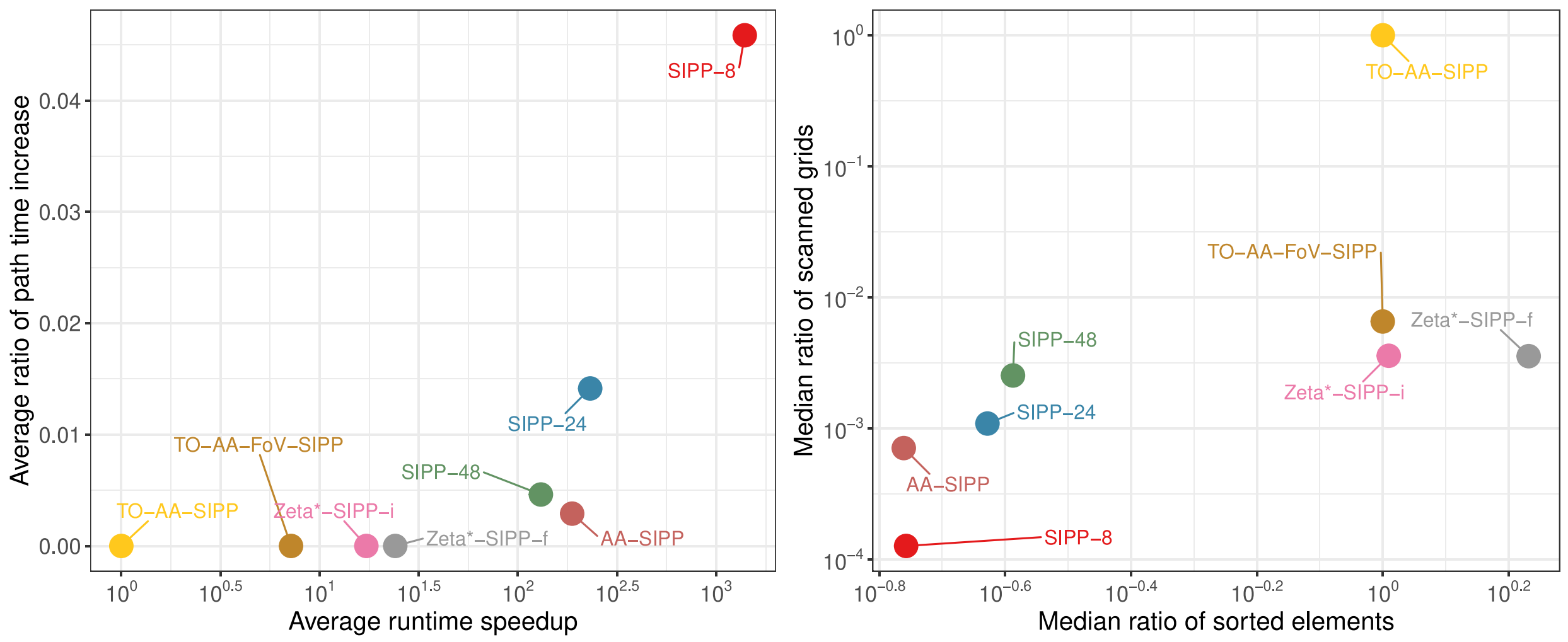

**Fig. 15.** Scatter plots of path time vs. runtime and scanned grids vs. sorted elements. All the metrics are normalized relative to TO-AA-SIPP. All axes, except for path time, are shown on a logarithmic scale due to the substantial differences between suboptimal and optimal planners. Median values are used in the right plot to reduce the influence of outliers, as most ratio values are very small.

While Zeta*-f clearly outperforms Zeta*-i, the performance gap between Zeta*-SIPP-f and Zeta*-SIPP-i is relatively small. This is because scan range pruning, as shown in Fig. 11, is only applicable in static environments. In such settings, forward scanning with range pruning can skip more unnecessary grid cells by leveraging taut-path rules and historical path information, since the scan is initialized from a closed node whose $g$-value is known and fixed. In contrast, inverted scanning starts from an open node, where no historical path is available.

Compared with the results reported in our previous conference paper [24], the runtimes of the optimal planners are substantially reduced, even though all experiments were conducted on the same machine. The performance difference between TO-AA-SIPP and the other optimal planners also becomes more pronounced. Two key factors contribute to this difference. First, the code for shadowcasting has been changed from an octant-based to a quadrant-based formulation and further optimized. Second, this work implements inverted expansion pruning to all the optimal planners, including TO-AA-SIPP, which reduces the list of potential parents for each open node.

Fig. 15 shows the overall benchmarking results of different path-planning algorithms relative to TO-AA-SIPP. It is evident that Zeta*-SIPP-f is slightly faster than Zeta*-SIPP-i, and both outperform TO-AA-SIPP by more than an order of magnitude, with around 24-fold and 17-fold speedups, respectively. Compared with TO-AA-FoV-SIPP, Zeta*-SIPP-f reduces the number of scanned grids through elliptical forward expansion, but this comes at the cost of increasing the number of sorted elements (i.e., the $scan$ list).

AA-SIPP runs at a speed between SIPP-24 and SIPP-48, but it can find better paths than both (with only about a 0.3% increase in path time compared to the optimal any-angle paths). Interestingly, although AA-SIPP scans more grids than SIPP-24 in the Maze, Ware, and Berlin maps (see Table 6), its overall median ratio of scanned grids relative to TO-AA-SIPP is slightly lower. This is because, in these three maps, TO-AA-SIPP scans an extremely large number of grids, resulting in a large denominator and reducing the ratio differences between AA-SIPP and SIPP-24.

## 7. Related work

Any-angle path planning addresses path-planning problems in continuous spaces. Although a graph is also used to discretize the environment, this approach relaxes the edge constraints and relies mainly on the graph vertices, allowing the discovery of shorter paths compared to traditional graph-based methods.

Theta* [5] is a well-known any-angle path-planning algorithm that has been widely used across various domains owing to its efficiency, simplicity, and generality. It leverages the triangular relationship among the current node, its parent, and its child, allowing the algorithm to bypass the current node and build a direct connection between the parent and child at each search step. Zeta* differs from Theta* in two ways: Zeta* straightens the search tree globally via *Field of View* (FoV) rather than locally via *Line of Sight* (LoS), and it employs any-angle (elliptical) rather than grid-by-grid (circular) forward expansion. These distinctions guarantee the optimality of Zeta*.

In this article, we propose using FoV in place of LoS to accelerate visibility checks. Similar improvements have also been investigated for Theta*. For example, Choi et al. [40] introduced two pruning strategies to improve the efficiency of LoS. The main idea is simple: since nodes around the "light source" may be scanned multiple times by LoS, this information can be reused to prevent redundant scans. This aligns with the original intention of FoV. The search node of Anya [13] can also be seen as FoV, which contributes to the algorithm's efficiency. In single-source any-angle path planning, wave-based algorithms like CWave [41] and Visibility-Based Marching (VBM) [42] utilize FoV to enhance performance as well, where the wave front is analogous to the scan depth in shadowcasting.

To guarantee any-angle optimality, we develop elliptical forward expansion. This technique was initially inspired by Informed Rapidly-exploring Random Trees* (Informed RRT*) [26] and Batch Informed Trees (BIT*) [27,43], as stated in our previous work [24]. Accelerated A* [44] also employs ellipses to improve path quality. Whereas Theta* checks only the current node's parent for a newly expanded node $n'$, Accelerated A* examines all closed nodes within a specified ellipse to identify the best parent for $n'$. The ellipse is defined with its foci at the start node $n_s$ and the node $n'$, with its major axis length set to $g(n')$. Accelerated A* also implements a dynamic adaptive expansion strategy. Instead of using a fixed number of neighbors (e.g., square neighborhoods), it expands nodes by identifying the maximum unblocked squares. With these two improvements, Accelerated A* is able to find shorter paths than Theta*. In the experiments of Šišlák et al. [44], it consistently produced the true shortest paths, although its optimality is not theoretically guaranteed.

Munoz and Rodriguez-Moreno [45] developed a new heuristic to improve Theta*, which incorporates heading changes. The goal of this heuristic is to prioritize the expansion of nodes that lie close to the direct line between the start and target points. Although no explicit ellipse is used, this idea is in line with elliptical forward expansion.

In uniform costmaps with static obstacles, true shortest paths must be taut [10]. This property has motivated the development of many algorithms based on obstacle corners, such as Strict Theta* [9], Anya [13], Polyanya [14], RayScan [17], R2 [18] and End Point Search (EPS) [19]. This type of problem is commonly known as the Euclidean Shortest Path Problem (ESPP) [19]. Zeta* presented in this article also incorporates this knowledge to restrict the search space and prune the scan range, significantly improving the algorithm performance.

From a practical perspective, the choice of planner for ESPP depends on whether the static obstacle configuration remains unchanged. For fixed static obstacles, methods such as Polyanya and EPS can be advantageous, since their preprocessing costs can be amortized over repeated path planning. In contrast, Anya and Zeta* are online grid-based planners that operate directly on the grid without requiring such map-dependent preprocessing, making them more suitable when the static obstacle configuration may change. There are other techniques for any-angle path planning, such as interpolation-based path cost calculation in Field D* [11], local distance database in Block A* [12], Subgoal Graphs [46], and $2^k$-neighborhoods for general grid-based path planning [37].

Although various any-angle path-planning algorithms have been proposed, only a few are capable of generating conflict-free paths in the presence of dynamic obstacles, such as Any-Angle Safe Interval Path Planning (AA-SIPP) [20], and Time-Optimal AA-SIPP (TO-AA-SIPP) [21]. The basic idea of AA-SIPP is similar to Theta*. It also seeks to straighten paths by checking whether the neighbors of the current node can be reached from the parent of the current node at a lower cost. However, this process is more complex than in Theta*, as the distance-based triangle inequality among the current node, its parent, and its neighbor (or child) is not applicable when waiting times are incorporated into the cost. Like Theta*, AA-SIPP is not guaranteed to find optimal any-angle paths.

To address this issue, TO-AA-SIPP was developed by checking whether the path to an open node can be straightened through any other closed node, rather than only its current parent. This idea is similar to Accelerated A*, but it does not restrict the set of closed nodes considered for straightening using an ellipse. TO-AA-SIPP has to examine visibility connections between all closed nodes and this open node. Moreover, TO-AA-SIPP inserts all nodes in the search space into *open* during initialization, and thus it could be very slow on large maps. Despite these shortcomings, to the best of our knowledge, TO-AA-SIPP is the first optimal planner for any-angle path planning in dynamic environments, which represents a meaningful step forward in this field. It has also been applied to develop the first optimal any-angle Multi-Agent Path Finding (MAPF) algorithm [23].

Our proposed algorithm, Zeta*-SIPP, is an improved version of TO-AA-SIPP. It implements elliptical forward expansion, similar to grid-by-grid forward expansion in A*, to reduce redundant nodes in the *open* list, and applies a field-of-view algorithm, shadowcasting, to accelerate visibility checks between nodes. These techniques are applicable to general grid-based path planning as well. In static environments, Zeta* is accordingly developed.

## 8. Conclusion

This article introduces two general techniques for optimal any-angle path planning: elliptical forward expansion and field of view. The first utilizes the geometric properties of ellipses to guarantee path optimality in any-angle path planning. The second accelerates online path smoothing by reducing redundant scanned grids. Zeta* and Zeta*-SIPP exemplify how these techniques can be applied. Zeta* is tailored for static environments. It is comparable to the state-of-the-art algorithm Anya, while being more easily extensible to other settings. Zeta*-SIPP demonstrates how to extend Zeta* to dynamic environments by integration with SIPP. Future studies can further explore their extensions to non-uniform costmaps and 3D scenarios.

In dynamic environments, optimal planners are significantly slower than their sub-optimal counterparts, since no taut-path constraint can be applied to restrict the search space. In this case, if real-time performance is the primary requirement for any-angle path planning, AA-SIPP is a suitable choice. If optimal solutions are desired—for example, when even small improvements in path quality can yield substantial savings in energy—Zeta*-SIPP is a recommended option. Benchmarking results show that Zeta*-SIPP-f is about 24 times faster than the current state-of-the-art optimal planner TO-AA-SIPP. Since Zeta*-SIPP is designed for dynamic environments, it also has the potential to serve as a low-level planner for optimal any-angle multi-agent path planning.

**CRediT authorship contribution statement**

**Yiyuan Zou:** Writing – original draft, Visualization, Validation, Software, Project administration, Methodology, Investigation, Funding acquisition, Formal analysis, Data curation, Conceptualization; **Clark Borst:** Writing – review & editing, Validation, Supervision, Resources, Project administration, Funding acquisition, Conceptualization.

## Declaration of competing interest

The authors declare that they have no known competing financial interests or personal relationships that could have appeared to influence the work reported in this paper.

## Acknowledgements

The authors would like to thank the China Scholarship Council for its financial support (No. 202106830036).

## Data availability

Code is available at https://github.com/yiyuanzou/zeta-sipp.

## References

[1] P.E. Hart, N.J. Nilsson, B. Raphael, A formal basis for the heuristic determination of minimum cost paths, IEEE Trans. Syst. Sci. Cybern. 4 (2) (1968) 100–107. https://doi.org/10.1109/TSSC.1968.300136
[2] Y. Zou, C. Borst, Algorithmic transparency in path planning: a visual approach to enhancing human understanding, Int. J. Hum.-Comput. Stud. 203 (2025) 103573. https://doi.org/10.1016/j.ijhcs.2025.103573
[3] A. Botea, M. Müller, J. Schaeffer, Near optimal hierarchical path-finding, J. Game Dev. 1 (1) (2004) 1–30.
[4] A. Nash, S. Koenig, Any-angle path planning, AI Mag. 34 (4) (2013) 85–107. https://doi.org/10.1609/aimag.v34i4.2512
[5] K. Daniel, A. Nash, S. Koenig, A. Felner, Theta*: any-angle path planning on grids, J. Artif. Intell. Res. 39 (2010) 533–579. https://doi.org/10.1613/jair.2994
[6] A. Nash, S. Koenig, M. Likhachev, Incremental Phi*: incremental any-angle path planning on grids, in: Proceedings of the 21st International Joint Conference on Artificial Intelligence, IJCAI'09, Morgan Kaufmann Publishers Inc., San Francisco, CA, USA, 2009, p. 1824–1830.
[7] S. Choi, W. Yu, Any-angle path planning on non-uniform costmaps, in: 2011 IEEE International Conference on Robotics and Automation, IEEE, 2011, pp. 5615–5621. https://doi.org/10.1109/ICRA.2011.5979769
[8] A. Nash, S. Koenig, C. Tovey, Lazy Theta*: any-angle path planning and path length analysis in 3D, in: Proceedings of the AAAI Conference on Artificial Intelligence, 24, 2010, pp. 147–154. https://doi.org/10.1609/aaai.v24i1.7566
[9] S. Oh, H.W. Leong, Strict Theta*: shorter motion path planning using taut paths, in: Proceedings of the International Conference on Automated Planning and Scheduling, 26, 2016, pp. 253–257. https://doi.org/10.1609/icaps.v26i1.13744
[10] J.S.B. Mitchell, D.M. Mount, C.H. Papadimitriou, The discrete geodesic problem, SIAM J. Comput. 16 (4) (1987) 647–668. https://doi.org/10.1137/0216045
[11] D. Ferguson, A. Stentz, Field D*: an interpolation-based path planner and replanner, in: Robotics Research: Results of the 12th International Symposium ISRR, Springer, 2007, pp. 239–253. https://doi.org/10.1007/978-3-540-48113-3_22
[12] P. Yap, N. Burch, R. Holte, J. Schaeffer, Block A*: database-driven search with applications in any-angle path-planning, in: Proceedings of the AAAI Conference on Artificial Intelligence, 25, 2011, pp. 120–125. https://doi.org/10.1609/aaai.v25i1.7813
[13] D.D. Harabor, A. Grastien, D. Öz, V. Aksakalli, Optimal any-angle pathfinding in practice, J. Artif. Intell. Res. 56 (2016) 89–118. https://doi.org/10.1613/jair.5007
[14] M.L. Cui, D.D. Harabor, A. Grastien, Compromise-free pathfinding on a navigation mesh, in: Proceedings of the 26th International Joint Conference on Artificial Intelligence, 2017, pp. 496–502.
[15] D.D. Harabor, New ideas for any-angle pathfinding, 2019, https://harabor.net/data/presentations/gdc2019.pdf. Accessed on Aug. 2nd, 2025.
[16] T. Uras, S. Koenig, An empirical comparison of any-angle path-planning algorithms, in: Proceedings of the International Symposium on Combinatorial Search, 6, 2015, pp. 206–210. https://doi.org/10.1609/socs.v6i1.18382
[17] R. Hechenberger, P.J. Stuckey, D. Harabor, P. Le Bodic, M.A. Cheema, Online computation of euclidean shortest paths in two dimensions, in: Proceedings of the International Conference on Automated Planning and Scheduling, 30, 2020, pp. 134–142. https://doi.org/10.1609/icaps.v30i1.6654
[18] Y.K. Lai, P. Vadakkepat, C. Xiang, R2: Optimal vector-based and any-angle 2D path planning with non-convex obstacles, Robot. Auton. Syst. 172 (2024) 104606. https://doi.org/10.1016/j.robot.2023.104606
[19] B. Shen, M.A. Cheema, D.D. Harabor, P.J. Stuckey, Fast optimal and bounded suboptimal Euclidean pathfinding, Artif. Intell. 302 (2022) 103624. https://doi.org/10.1016/j.artint.2021.103624
[20] K. Yakovlev, A. Andreychuk, Any-angle pathfinding for multiple agents based on SIPP algorithm, in: Proceedings of the International Conference on Automated Planning and Scheduling, 27, 2017, pp. 586–594. https://doi.org/10.1609/icaps.v27i1.13856
[21] K. Yakovlev, A. Andreychuk, Towards time-optimal any-angle path planning with dynamic obstacles, in: Proceedings of the International Conference on Automated Planning and Scheduling, 31, 2021, pp. 405–414. https://doi.org/10.1609/icaps.v31i1.15986
[22] M. Phillips, M. Likhachev, SIPP: safe interval path planning for dynamic environments, in: 2011 IEEE International Conference on Robotics and Automation, 2011, pp. 5628–5635. https://doi.org/10.1109/ICRA.2011.5980306
[23] K. Yakovlev, A. Andreychuk, R. Stern, Optimal and bounded suboptimal any-angle multi-agent pathfinding, in: 2024 IEEE/RSJ International Conference on Intelligent Robots and Systems (IROS), IEEE, 2024, pp. 7996–8001. https://doi.org/10.1109/IROS58592.2024.10801691
[24] Y. Zou, C. Borst, Zeta*-SIPP: improved time-optimal any-angle safe-interval path planning, in: Proceedings of the 33rd International Joint Conference on Artificial Intelligence, 2024, pp. 6823–6830. https://doi.org/10.24963/ijcai.2024/754
[25] J.P. Bailey, A. Nash, C.A. Tovey, S. Koenig, Path-length analysis for grid-based path planning, Artif. Intell. 301 (2021) 103560. https://doi.org/10.1016/j.artint.2021.103560
[26] J.D. Gammell, S.S. Srinivasa, T.D. Barfoot, Informed RRT*: optimal sampling-based path planning focused via direct sampling of an admissible ellipsoidal heuristic, in: 2014 IEEE/RSJ International Conference on Intelligent Robots and Systems, 2014, pp. 2997–3004. https://doi.org/10.1109/IROS.2014.6942976
[27] J.D. Gammell, S.S. Srinivasa, T.D. Barfoot, Batch informed trees (BIT*): sampling-based optimal planning via the heuristically guided search of implicit random geometric graphs, in: 2015 IEEE International Conference on Robotics and Automation (ICRA), 2015, pp. 3067–3074. https://doi.org/10.1109/ICRA.2015.7139620
[28] S.J. Russell, P. Norvig, Artificial Intelligence: A Modern Approach, Pearson Education, Inc., 2010.
[29] J.E. Bresenham, Algorithm for computer control of a digital plotter, IBM Syst. J. 4 (1) (1965) 25–30. https://doi.org/10.1147/sj.41.0025
[30] X. Wu, An efficient antialiasing technique, Acm Siggr. Comput. Graph. 25 (4) (1991) 143–152.
[31] B. Bergström, FOV using recursive shadowcasting, 2001, (http://www.roguebasin.com/index.php?title=FOV_using_recursive_shadowcasting). Accessed on Aug. 2nd, 2025.
[32] A. Milazzo, Roguelike vision algorithms, 2014, (https://www.adammil.net/blog/v125_Roguelike_Vision_Algorithms.html). Accessed on Aug. 2nd, 2025.
[33] E.R.M. Debenham, R. Solis-Oba, Efficient field of vision algorithms for large 2D grids, Int. J. Comput. Sci. Inf. Technol. 13 (2021) 1–20. https://doi.org/10.5121/ijcsit.2021.13101
[34] X. Xu, PathFinding.js, 2026 (https://github.com/qiao/PathFinding.js). Accessed on May 23rd, 2026.
[35] N. Sturtevant, Benchmarks for grid-based pathfinding, Trans. Comput. Intell. AI Games 4 (2) (2012) 144–148. https://doi.org/10.1109/TCIAIG.2012.2197681

[36] D. Harabor, R. Hechenberger, T. Jahn, Benchmarks for pathfinding search: iron harvest, in: Proceedings of the International Symposium on Combinatorial Search, 15, 2022, pp. 218–222.

[37] N. Rivera, C. Hernández, N. Hormazábal, J.A. Baier, The $2^k$ neighborhoods for grid path planning, J. Artif. Intell. Res. 67 (2020) 81–113. https://doi.org/10.1613/jair.1.11383

[38] D. Silver, Cooperative pathfinding, in: Proceedings of the AAAI Conference on Artificial Intelligence and Interactive Digital Entertainment, 1, 2005, pp. 117–122. https://doi.org/10.1609/aiide.v1i1.18726

[39] R. Stern, N.R. Sturtevant, A. Felner, S. Koenig, H. Ma, T.T. Walker, J. Li, D. Atzmon, L. Cohen, T.K.S. Kumar, E. Boyarski, R. Bartak, Multi-agent pathfinding: definitions, variants, and benchmarks, Symp. Comb. Search (2019) 151–158. https://doi.org/10.1609/socs.v10i1.18510

[40] S. Choi, J.-Y. Lee, W. Yu, Fast any-angle path planning on grid maps with non-collision pruning, in: 2010 IEEE International Conference on Robotics and Biomimetics, IEEE, 2010, pp. 1051–1056. https://doi.org/10.1109/ROBIO.2010.5723473

[41] D.A. Sinyukov, T. Padir, CWave: Theory and practice of a fast single-source any-angle path planning algorithm, Robotica 38 (2) (2020) 207–234. https://doi.org/10.1017/S0263574719000560

[42] I. Ibrahim, J. Gillis, W. Decré, J. Swevers, Exact wavefront propagation for globally optimal one-to-all path planning on 2D cartesian grids, IEEE Robot. Autom. Lett. (2024). https://doi.org/10.1109/LRA.2024.3460409

[43] J.D. Gammell, T.D. Barfoot, S.S. Srinivasa, Batch informed trees (BIT*): informed asymptotically optimal anytime search, Int. J. Robot. Res. 39 (5) (2020) 543–567. https://doi.org/10.1177/0278364919890396

[44] D. Šišlák, P. Volf, M. Pechoucek, Accelerated A* trajectory planning: grid-based path planning comparison, in: Proceedings of the 19th International Conference on Automated Planning & Scheduling (ICAPS), Citeseer, 2009, pp. 74–81.

[45] P. Munoz, M. Rodriguez-Moreno, Improving efficiency in any-angle path-planning algorithms, in: 2012 6th IEEE International Conference Intelligent Systems, IEEE, 2012, pp. 213–218. https://doi.org/10.1109/IS.2012.6335138

[46] T. Uras, S. Koenig, Speeding-up any-angle path-planning on grids, in: Proceedings of the International Conference on Automated Planning and Scheduling, 25, 2015, pp. 234–238. https://doi.org/10.1609/icaps.v25i1.13724